\documentclass[journal]{IEEEtran}
\usepackage{amsmath,amsfonts}
\usepackage{algorithmic}
\usepackage{algorithm}
\usepackage{tabularx}
\usepackage{multirow}
\usepackage{makecell}
\usepackage{threeparttable}
\usepackage{booktabs}
\usepackage{amssymb}
\usepackage{array}
\usepackage[caption=false,font=normalsize,labelfont=sf,textfont=sf]{subfig}
\usepackage{textcomp}
\usepackage{stfloats}
\usepackage{url}
\usepackage{verbatim}
\usepackage{graphicx}
\usepackage{cite}
\begin{document}

\title{STLQ: Spatio-Topological Linkage Quantization for Approximate Nearest Neighbor Search}

\author{Kuizhi Liu,
        Junying Zhang,
        Xindi Ma,~\IEEEmembership{Member,~IEEE,}
        Xiangyu Wang,~\IEEEmembership{Member,~IEEE,}
        Qi Jiang,
        Sheng Gao,
        Chen Gong,
        and Jianfeng Ma,~\IEEEmembership{Member,~IEEE}
\thanks{K. Liu, J. Zhang, X. Wang, Q. Jiang, and J. Ma are with the State Key Laboratory of Integrated Service Networks (ISN), Xidian University, Xi'an, China. They are also with the School of Cyber Engineering, Xidian University, Xi'an, China. \protect
E-mail: ighxiy@163.com, jyzhang1990@gmail.com, xywang\_xidian@163.com, jiangqixdu@gmail.com, jfma@mail.xidian.edu.cn.}
\thanks{X. Ma is with the School of Cyber Engineering, Xidian University, Xi'an, China. He is also with the Xidian University Qingdao Computing Technology Research Institute, Qingdao, China. \protect E-mail: xdma1989@gmail.com. }
\thanks{S. Gao is with the School of Information, Central University of Finance and Economics, Beijing, China. \protect E-mail: sgao@cufe.edu.cn.}
\thanks{Corresponding author: Xindi Ma.}
\thanks{This work has been submitted to the IEEE for possible publication.
Copyright may be transferred without notice, after which this version
may no longer be accessible.}}

\markboth{Journal of \LaTeX\ Class Files,~Vol.~14, No.~8, August~2021}%
{Shell \MakeLowercase{\textit{et al.}}: A Sample Article Using IEEEtran.cls for IEEE Journals}


\maketitle

\begin{abstract}
Vector quantization is widely used for large-scale data compression and approximate nearest neighbor search (ANNS). Multi-codebook quantization (MCQ) is a practical paradigm that represents database vectors with compact codes and supports efficient retrieval through asymmetric distance computation (ADC). However, existing shallow MCQs encode vectors independently within each inverted file (IVF) cluster, without exploiting local references. Therefore, we propose Spatio-Topological Linkage Quantization (STLQ), a topological shallow MCQ method that organizes each IVF cluster as a directed-forest with compact codes and scaling coefficients. Its basic encoding phase first quantizes vectors independently, while the linked encoding phase re-encodes relative residuals with respect to selected reconstructed parents, reusing local reconstruction information to obtain finer representations. The corresponding linked-ADC preserves linear scanning complexity. STLQ further compresses the scaling coefficients and directed-forest structure to control the additional topology overhead. Extensive experiments show that STLQ achieves higher recall than the evaluated shallow MCQs at comparable bit budgets, although at a higher offline construction cost. Meanwhile, it has a competitive recall-QPS tradeoff against deep MCQs. The code is available at https://github.com/ighxiy/STLQ.
\end{abstract}


\section{Introduction}
\label{sec_introduction}
\IEEEPARstart{V}{ector} retrieval is a fundamental component in vector databases, recommendation systems, and retrieval-augmented generation. These systems need to balance accuracy, query efficiency, and storage cost. However, full-precision retrieval incurs large memory and computational costs. Vector quantization (VQ) uses compact representations to address it, and is widely used for approximate nearest neighbor search (ANNS) in similarity search \cite{zhang2023query} and recommendation \cite{zheng2024adapting} on billion-scale data. VQ is also used for image generation\cite{zhang2023regularized} and large language model compression \cite{pmlr-v235-egiazarian24a,li2025commvq}.

Product Quantization (PQ) \cite{jegou2010product} established the practical multi-codebook quantization (MCQ) paradigm, which represents vectors with compact codes and estimates their distances to a query through Asymmetric Distance Computation (ADC). Subsequent methods improve the representation capacity through different codebook structures and encoding strategies. However, in conventional shallow MCQs, vectors within the same inverted file (IVF) cluster are still encoded independently. Therefore, the local spatial correlation within each IVF cluster remains underexplored.

The encoding objective function of MCQ is to minimize the Mean Squared Error (MSE) between original and quantized vectors, which is termed as quantization error. Based on residual-style encoding, we observe that vectors farther from the IVF centroid tend to exhibit larger quantization errors, as their residuals are generally larger and more directionally deviated. This motivates us to use reconstructed vectors within the same cluster as finer references, so that a vector can encode a smaller relative residual instead of always referring to the IVF centroid. However, the nearest vector is not necessarily a good reference, since the reference itself is also quantized. Therefore, candidate references should first satisfy spatial constraints, and the final reference should be selected by the reconstruction MSE after residual quantization rather than geometric distance.

Motivated by these observations, we propose STLQ, a Spatio-Topological Linkage Quantization method. Instead of encoding every vector independently, STLQ allows a vector to reuse the reconstruction of another vector in the same IVF cluster and encode only the remaining relative residual, which is a parent-conditioned residual encoding mechanism. A linkage is accepted only when it reduces the reconstruction MSE compared with independent encoding. The accepted parent-child relationships are organized as a directed-forest to keep these dependencies simple and support efficient queries. This forest is an encoding dependency structure rather than a navigation graph. During retrieval, linked-ADC reuses the accumulated parent score along the forest with linear scanning complexity.

Extensive experiments are conducted to evaluate STLQ in terms of recall, queries per second (QPS), bit budget, storage cost, and index building time. STLQ achieves higher recall than the evaluated shallow MCQ methods at comparable storage budgets, while retaining a high QPS. Nevertheless, STLQ uses a higher offline construction cost to form the directed-forest. Compared with deep MCQs, STLQ does not uniformly achieve the highest recall, but provides a competitive recall-QPS tradeoff. Compared with data-oblivious quantization, STLQ achieves a better rate-recall tradeoff, while data-oblivious methods provide higher QPS, and substantially lower index construction cost.

In summary, the contributions are as follows:

$\bullet$ We propose STLQ, a novel parent-conditioned residual encoding MCQ method, where a vector reuses a reconstructed parent and quantizes only the remaining relative residual, with the parent selected by MSE rather than geometric distance.

$\bullet$ We design a cluster-adaptive directed-forest structure with outward-expanding and virtual-relay strategies for different cluster distributions. 

$\bullet$ We develop linked-ADC and compact index representations that preserve linear scanning complexity while controlling the additional storage overhead.

\section{Related Work}
We review VQ methods for ANNS according to their codebook structures and encoding strategy. We mainly discuss MCQ paradigm. Moreover, we mention a different paradigm, data-oblivious quantization, only as a complement.

\label{lab_related_work}
\subsection{Shallow MCQ}
\subsubsection{Cartesian Product Structure}
PQ \cite{jegou2010product} established the practical MCQ methods and lookup-table-based ADC scanning for large-scale ANNS. It quantizes vector subspaces separately with a Cartesian product structure. This structure provides high query efficiency, but the independence between subspaces limits representation quality. Optimized Product Quantization (OPQ) \cite{ge2013optimized} addresses this problem by learning an orthogonal rotation through the orthogonal Procrustes problem (OPP), which balances variance across subspaces. Locally Optimized Product Quantization (LOPQ) \cite{kalantidis2014locally} further introduces cluster-specific transformations. PQ-based methods are attractive for large-scale retrieval because of their higher QPS. Further optimization and hardware acceleration of PQ have also been studied \cite{johnson2019billion,liu2024juno}.

\subsubsection{Additive Structure}
Residual Vector Quantization (RVQ) \cite{chen2010approximate} uses a hierarchical structure to encode residuals with full length codewords. Later, Additive Quantization (AQ) \cite{babenko2014additive} uses parallel codebooks to further improve the accuracy, which allows codewords from different codebooks to interact freely. However, such a Markov Random Field (MRF) procedure is NP-hard to optimize, making the beam search expensive in AQ. This problem is alleviated in Orthogonal Tree Quantization (OTQ) \cite{babenko2015tree}. Composite Quantization (CQ) \cite{zhang2014composite} introduces constraints on cross interactions of codebooks to speed up the retrieval. Competitive Quantization (CompQ) \cite{ozan2016competitive} jointly optimizes hierarchical additive codebooks by stochastic gradient descent (SGD). However, the large epochs of sequential SGD restrict its application. Local Search Quantization (LSQ) and LSQ++ \cite{martinez2016revisiting, martinez2018lsq++} improve AQ with Iterated Local Search (ILS) \cite{lourencco2003iterated} and Iterated Conditional Modes (ICM) \cite{besag1986statistical}, which provide practical local optimization of code assignments without exhaustive search over the combined codebook space. STLQ is closest to this family in using explicit additive codebooks and an LSQ-style optimization backbone. The main difference is that these shallow MCQs represent database vectors independently, whereas STLQ additionally allows a vector to reuse another reconstructed vector in the same IVF cluster as its encoding reference. In addition, STLQ introduces scaling coefficients to adapt variations in residual magnitude.

\subsection{Deep MCQ} Deep MCQs increase representation capacity by using learned nonlinear encoders, decoders, or adaptive codewords. Unsupervised Neural Quantization (UNQ) \cite{morozov2019unsupervised} learns a neural compact representation for ANNS, while DR-UNQ \cite{noh2023disentangled} further improves the accuracy by introducing IVF priors to both the encoder and decoder. QINCo \cite{huijben2024residual} uses implicit neural codebooks, where the codeword at each stage is conditioned on previously selected codes, producing adaptive codewords and lower MSE, but with substantial training and retrieval costs. QINCo2 \cite{vallaeys2025qinco2} further improves this design through stronger implicit codebooks, achieving higher representation quality at increased encoding complexity. It is noted that QINCo family also uses a conditional representation. However, QINCo conditions later codewords on the previously selected codes of the same vector, whereas STLQ conditions a linked representation on the reconstruction of another vector in the same IVF cluster.

\subsection{Data-Oblivious Quantization}
RaBitQ \cite{gao2024rabitq,gao2025practical} and TurboQuant \cite{zandieh2025turboquant} follow a data-oblivious quantization paradigm, rather than the data-learned MCQ paradigm. They construct quantizers without learning dataset-specific codebooks, and provide explicit distortion guarantees based on random projections and their quantized variants \cite{johnson1984extensions,jacques2015quantized}. Consequently, they usually achieve higher QPS than MCQs, and are suitable for online services. However, their recall may be lower than that of some MCQs without reranking, and the storage model depends on vector dimensionality rather than a fixed MCQ code length. STLQ instead belongs to the MCQ with a fixed rate, higher offline construction cost, and higher recall. We therefore consider data-oblivious quantizers as a complement to learned MCQ rather than the closest paradigm of STLQ.

\section{Preliminaries}
\subsection{Problem Formulation and Design Goals}
Let $\{\mathbf{x}\}_{i=1}^{N} \subset \mathbb{R}^{d}$ be $N$ database vectors. STLQ uses IVF as a coarse quantizer, and encodes $\mathbf{x}$ cluster-wise into compact codes and coefficients under a fixed storage budget. Let $\mathbf{q}\in\mathbb{R}^{d}$ denote a query vector, STLQ uses linked-ADC to search the compressed database, without accessing the original database vectors. During encoding, STLQ minimizes MSE between the original and reconstructed vectors as a surrogate objective, aiming to improve recall. MSE is therefore used to optimize the vector representation, and select parent relationships, whereas retrieval quality is finally evaluated by recall and QPS. We report both the payload bit budget $\beta_p$, which measures the vector representation itself, and the total bit budget $\beta_t$, which additionally accounts for structural information required by retrieval. Offline cost includes training time and database encoding time. The design goals of STLQ are to achieve high retrieval accuracy while keeping the query efficiency under the fixed storage budget.

\subsection{Representation of STLQ}
\label{lab_stlq_representation}
Table \ref{tab_symbol} summarizes the symbols used in STLQ. STLQ contains two encoding phases. \textbf{Basic Encoding} independently represents each vector using the \textbf{Root Codebooks} $\mathbf{C}^r=\{\mathbf{C}_1^r,\mathbf{C}_2^r,\ldots,\mathbf{C}_M^r \mid \mathbf{C}_1^r \in \mathbb{R}^{d\times h_1^r}, \mathbf{C}_i^r \in \mathbb{R}^{d\times h}, i\neq 1\}$. The first root codebook also serves as the integrated IVF coarse quantizer rather than using an external IVF index. \textbf{Linked Encoding} allows a vector to reuse the reconstruction of another vector in the same IVF cluster, and encode only the remaining relative residual using the \textbf{Residual Codebooks} $\mathbf{C}^o=\{\mathbf{C}_1^o,\mathbf{C}_2^o,\ldots,\mathbf{C}_M^o \mid \mathbf{C}_i^o \in \mathbb{R}^{d\times h}\}$. 

Since IVF clusters have different spatial distributions, STLQ estimates their compactness from the cluster-wise MSE, as a broader angular distribution relative to the IVF centroid usually leads to higher MSE. The clusters are then divided into \textbf{Compact Cluster} and \textbf{Dispersed Cluster} according to MSE, where the clusters with the top $\gamma_d$ fraction by MSE are marked as dispersed ones.

For node $i$, $p(i)=\varnothing$, equivalently $D_i=0$, denotes a \textbf{Root Node}; $D_i\geq 1$ denotes a \textbf{Linked Node}. A directed edge $p(i)\rightarrow i$ indicates that node $i$ reuses the reconstruction $\hat{\mathbf{x}}_{p(i)}$ of its parent, and encodes the relative residual $\epsilon_{i|p}=\mathbf{x}_i-\hat{\mathbf{x}}_{p(i)}$. The accepted parent-child relationships satisfy the \textbf{Directed-Forest} constraint: each node has at most one parent and the dependency structure is acyclic. A \textbf{Linkage} is a continuous path from a root node to linked nodes.

For each vector, topology construction first provides a candidate parent set $\tilde{p}(i)$. The final parent $p(i)$ is not determined by geometric distance. Each candidate is evaluated after residual quantization, and the candidate with the lowest reconstruction MSE is selected only if it improves over the basic encoding result. Otherwise, the vector remains a root node. The details are introduced in Section \ref{lab_construction_method}.

The reconstruction of node $i$ can be written as:
\begin{equation}
\hat{\mathbf{x}}_{i} =
    \begin{cases}
      \sum\limits_{m=1}^{M}
      \alpha_{i,m}\mathbf{C}_{m}^{r}\mathbf{b}_{i,m},& D_i=0,
      \\
      \hat{\mathbf{x}}_{p(i)} + \sum\limits_{m=1}^{M} \alpha_{i,m}\mathbf{C}_{m}^{o}\mathbf{b}_{i,m},& D_i \ge 1,
    \end{cases}
\label{fun_x_reconstruction}
\end{equation}
where $\mathbf{b}_{i,m}$ is a one-hot index that selects the codeword $\mathbf{c}_{i,m}$.

\begin{table}[!htb]
\centering
\caption{Symbols in STLQ}
\label{tab_symbol}
\begin{tabular}{l l}
\toprule
\textbf{Symbols} & \textbf{Description} \\
\midrule
$\{\mathbf{x}\}_{i=1}^{N}$; $\{\hat{\mathbf{x}}\}_{i=1}^{N}$& $N$ database vectors $\mathbf{x} \in \mathbb{R}^{d}$; Reconstructed ones\\

$\mathbf{e}$ & Quantization error $\mathbf{e}=\mathbf{x}-\hat{\mathbf{x}}$\\

$\mathbf{q}$ & Query vector $\mathbf{q} \in \mathbb{R}^{d}$\\

$\mathbf{R}$ & Orthogonal rotation matrix for $\hat{\mathbf{x}}$\\

$n$ & Number of vectors within an IVF cluster\\

$M$; $m$ & $M$ codebooks form a set, each is indexed by $m$\\

$\mathbf{C}^r$; $\mathbf{C}^o$ & A set of root codebooks, and residual codebooks\\

$\mathbf{c}_m$; $\mathbf{C}_m$ & The $m$-th selected codeword, and codebook\\

$h_1^r$& Codeword size in $\mathbf{C}_1^r$, can be large\\

$h$ & Codeword size of other layers, typically $h=256$\\

$\mathbf{b}_i;\mathbf{B}$ & Codeword indices of $\mathbf{x}_i$; Index matrix for $\{\hat{\mathbf{x}}\}_n$ \\

$\alpha_{i}$; $\hat{\alpha}_{i}$ & Scaling coefficient per $\mathbf{b}_i$; Quantized $\alpha_{i}$ for $\{\hat{\mathbf{x}}\}_{N}^{d}$\\

$p(i)$; $\tilde{p}(i)$ & Final parent of node $i$; Candidate parent set\\

$\mathcal{P}$ & Parent array in level order within an IVF cluster\\

$\mathcal{F}_i$ & Directed-forest of $i$-th IVF cluster stored as $\mathcal{P}$\\

$\epsilon_{i|p}$ & Relative residual $\epsilon_{i|p}=\mathbf{x}_i-\hat{\mathbf{x}}_{p(i)}$\\

$D_i; D_{max}$ & Topological depth of node $i$; Depth max cap \\

$\mathbb{E}$; $\mathbb{S}$ & MSE; Cosine similarity \\

$w$ & Beam width in basic encoding phase\\

$\mathcal{L}$; $z$ & Topology layer number; Depth-window $z \leq \mathcal{L}$\\

$\gamma_{l}$ & Linked ratio: linked nodes size divided by $N$ \\

$\gamma_{v}$ & Virtual ratio: virtual nodes size divided by $N$\\

$\gamma_{d}$ & Dispersed cluster ratio, whose denominator is $h_1^r$\\

$\beta_{p}; \beta_{t}$ & Payload bit budget; Total bit budget\\
\bottomrule
\end{tabular}
\end{table}

\subsection{Objective Function}
\label{lab_objective_function}
\subsubsection{Encoding Objective}
The training objective is to minimize the reconstruction MSE by alternately updating the code assignments, scaling coefficients, codebooks, topology, and rotation matrix:
\begin{equation}
\min_{\alpha,\mathbf{b},\mathbf{c},\mathcal{F},\mathbf{R}}(\mathbb{E}) = \min \frac{1}{N} \sum\nolimits_{i=1}^N \| \mathbf{R}\mathbf{x}_{i}-\hat{\mathbf{x}}_{i} \|^2.
\label{fun_objective_function}
\end{equation}

\subsubsection{ADC Retrieval Objective}
ADC uses $\hat{\mathbf{x}}$ instead of $\mathbf{x}$ to calculate query distances through lookup tables (LUTs) and stored pre-computed norm terms. Combining with Eq. \ref{fun_x_reconstruction}, define $\mathbf{q}'=\mathbf{Rq}$, the query ranking function is:
\begin{equation}
    \mathbf{q}(\mathbf{x})=\arg\min_{\hat{\mathbf{x}}}\|\mathbf{q}' - \hat{\mathbf{x}} \|^2 = \arg\min_{\hat{\mathbf{x}}} (-2{\mathbf{q}'}^T\hat{\mathbf{x}} + \| \hat{\mathbf{x}} \|^2).
    \label{fun_adc_score}
\end{equation}
The directed-forest representation changes how $\hat{\mathbf{x}}$ is accumulated, but does not change the ADC ranking objective. Using LUT, the pre-computed norm term, and the level-order parent array $\mathcal{P}$, linked-ADC preserves the same $O(n)$ scanning complexity as additive ADC.

\subsubsection{MSE and Ranking Error}
MSE is a proxy metric of quantization quality, and a lower MSE generally leads to higher recall over a broad distortion range. However, within a narrow MSE range, the relationship can be non-monotonic because recall is determined by pairwise ranking. Suppose $\mathbf{x}_1$ is the answer and $\mathbf{x}_2$ is a competitor. Let $\Delta_{e}$ and $\Delta_{adc}$ denote their exact and ADC differential scores, respectively:
\begin{align}
    &\Delta_{e} \!=\! \|\mathbf{q}' \!-\! \mathbf{x}_2 \|^2 \!-\! \|\mathbf{q}' \!-\! \mathbf{x}_1 \|^2, 
    \Delta_{adc} \!=\! \|\mathbf{q}' \!-\! \hat{\mathbf{x}}_2 \|^2 \!-\! \|\mathbf{q}' \!-\! \hat{\mathbf{x}}_1 \|^2 ,\nonumber\\
    &\Delta_{adc} = \Delta_{e} + 2\mathbf{q}'\cdot(\mathbf{e}_2 - \mathbf{e}_1) \nonumber\\
    &\hphantom{\Delta_{adc} = }+ (\| \hat{\mathbf{x}}_2 \|^2 - \| \mathbf{x}_2 \|^2) - (\| \hat{\mathbf{x}}_1 \|^2 - \| \mathbf{x}_1 \|^2).
    \label{fun_diff_err}
\end{align}
Eq. \ref{fun_diff_err} decomposes the ranking error introduced by quantization into an inner-product term and a norm term. Therefore, reconstruction MSE alone does not completely determine recall. We analyze these ADC error terms empirically in the evaluation section to explain the retrieval behavior of STLQ.

\section{Construction of STLQ}
\label{lab_construction_method}
\begin{figure*}[!htb]
  \centering
  \includegraphics[width=1\textwidth]{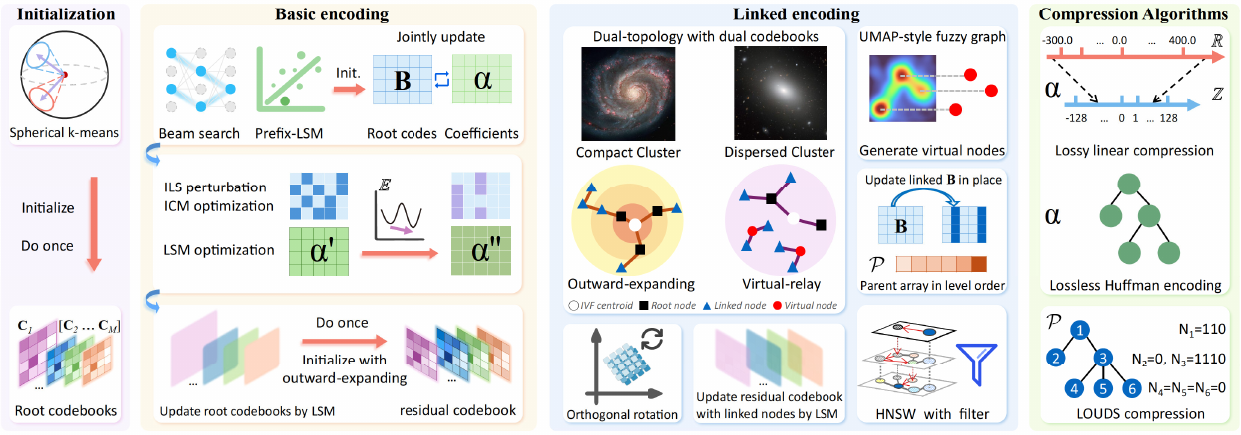}
  \caption{STLQ technical overview. Basic encoding generates independent codes and coefficients, while linked encoding accepts parent-dependent representations only when they reduce reconstruction MSE. Compact and dispersed clusters use outward-expanding and virtual-relay topologies, respectively. Training alternates between the two encoding phases. After base set encoding, coefficients and forest topology are compressed for storage. Spiral and elliptical galaxies hint the cluster structure entropy and topologies. Credit: NASA, ESA, S. Beckwith (STScI) and the Hubble Heritage Team (STScI/AURA). [Public domain], via NASA (https://www.nasa.gov/).}
  \label{fig_technical_overview}
\end{figure*}

Given the representation and objective defined in Section \ref{lab_stlq_representation} and \ref{lab_objective_function}, STLQ jointly optimizes the forest topology, code assignments, scaling coefficients, and dual codebooks through two encoding phases. Basic encoding provides independent fallback representations, while linked encoding evaluates feasible reconstructed parents that are accepted only if MSE decreases. Training alternates between these phases, codebook updates, and orthogonal rotation. Compression controls the additional storage overhead, while linked-ADC supports linear scanning. Fig. \ref{fig_technical_overview} and Algorithm \ref{algori_STLQ} summarize the framework and training procedure. The detailed mathematical derivations are provided in the Appendix.

These components address different consequences of introducing cross-vector dependency. Basic encoding provides a fallback representation, the dual topology constrains feasible dependencies to a directed-forest in linked encoding, and the MSE criterion admits only beneficial linkages. Compression and linked-ADC then make the forest representation compatible with the storage, and enable linear retrieval.

\subsection{Basic Encoding: Establishing an Independent Baseline}
Basic encoding generates the initial code assignments, scaling coefficients, and MSE for each vector by root codebooks $\mathbf{C}^{r}$. These representations are the fallback for linked encoding, where a linked representation replaces the basic one only if it reduces MSE

\begin{algorithm}[!htb]
  \renewcommand{\algorithmicrequire}{\textbf{Input:}}
  \renewcommand{\algorithmicensure}{\textbf{Output:}}
  \caption{\texttt{STLQ Training Algorithms}.}\label{algori_STLQ}
  \begin{algorithmic}[1]
  \REQUIRE train set $\mathbf{X}$, initial orthogonal matrix $\mathbf{R}_0=\mathbf{I}$, number of iterations $T$;
  \ENSURE dual codebooks $\mathbf{C}^r$ and $\mathbf{C}^o$, $\mathbf{R}_T$ after final iteration;
  \STATE Initialize $\mathbf{C}^r$ by spherical k-means, orthogonal projection;
  \STATE Initialize ($\mathbf{B}$, $\alpha$) by $\mathbf{C}^r$, beam search, and prefix-LSM;
  \STATE Optimize ($\mathbf{B}$, $\alpha$) by LSM, ILS, and ICM;
  \STATE Update $\mathbf{C}^r$ by LSM on all nodes;
  \STATE Update ($\mathbf{B}$, $\alpha$) by $\mathbf{C}^r$ and outward-expanding topology;
  \STATE Initialize $\mathbf{C}^o$ by LSM on linked nodes;
  \FOR {$t = 1$ to $T$}
    \STATE Rotate dataset $\mathbf{X}'=\mathbf{R}_{t-1} \mathbf{X}$ for current iteration;
    \STATE Initialize ($\mathbf{B}_{t}$, $\alpha_{t}$) by $\mathbf{C}_{t}^r$, beam search, and prefix-LSM;
    \STATE Optimize ($\mathbf{B}_{t}$, $\alpha_{t}$) by LSM, ILS, and ICM;
    \STATE Update $\mathbf{C}_{t}^r$ by LSM on all nodes;
    \STATE Update ($\mathbf{B}_{t}$, $\alpha_{t}$) by $\mathbf{C}_{t}^o$ and dual topologies;
    \STATE Update $\mathbf{C}_{t}^o$ by LSM on linked nodes;
    \STATE Update $\mathbf{R}_{t}$ by OPP;
  \ENDFOR
  \STATE \textbf{return} $\mathbf{C}^r$, $\mathbf{C}^o$ and $\mathbf{R}_T$;
  \end{algorithmic}
\end{algorithm}

\subsubsection{Root Codebook Initialization}
$\mathbf{C}^{r}$ is initialized layer by layer in RVQ style. The train set $\mathbf{x}$ is clustered by annealing spherical k-means to obtain the first layer codewords that simultaneously serve as the integrated IVF coarse quantizer. Consequently, $h_1^r$ can be much larger than the sizes of the remaining layers. Then $\mathbf{x}$ are orthogonally projected to the nearest unit codeword to get the coefficients and residuals that are the input to the next layer.

\subsubsection{Beam Search with Prefix-LSM}
$\mathbf{C}^{r}$ is then used to encode $\mathbf{x}$ by beam search with prefix-LSM. Beam search retains $w$ candidate code prefixes rather than a single one. Because a new state changes the best $\alpha$, beam search and LSM must update $\alpha$ jointly during each layer. For a prefix $\mathbf{C}_k^r=[\mathbf{c}_1^r,\ldots,\mathbf{c}_k^r]\in\mathbb{R}^{d\times k}$, the state energy is:
\begin{equation}
    \min_{\alpha}(\mathbb{E}_k)= \min_{\alpha}\|\mathbf{x} - \mathbf{C}_k^r\alpha_k\|^2,\quad \alpha_k\in\mathbb{R}^{k\times 1}.
    \label{fun_prelsm_coeff_obj}
\end{equation}
Solving this least squares problem for each expansion is expensive. Prefix least squares method (Prefix-LSM) instead reuses the factorization of $\mathbf{G}_k={\mathbf{C}_k^r}^{T}\mathbf{C}_k^r$. For a candidate codeword $\mathbf{c}_n^r$, define $\mathbf{U}={\mathbf{C}_k^r}^{T}\mathbf{c}_n^r$, $g=\|\mathbf{c}_n^r\|^2$, and $v={\mathbf{c}_n^r}^{T}\mathbf{x}$. The new coefficient $\alpha_n$ and the updated prefix coefficients $\tilde{\alpha}_k$ are:
\begin{align}
    &\alpha_n = s^{-1}(v-\mathbf{U}^{T}\alpha_k),\quad \tilde{\alpha}_k=\alpha_k - \alpha_n\mathbf{G}_k^{-1}\mathbf{U},\nonumber\\ 
    &s = g-\mathbf{U}^{T}\mathbf{G}_k^{-1}\mathbf{U}.
    \label{fun_solution_prefix_lsm}
\end{align}
Thus an extension refits the whole prefix, rather than merely appending a projection coefficient. After solving the best $\alpha$ for the current state, beam search selects the top-$w$ candidates by energy:
\begin{equation}
    \min_{\alpha}(\mathbb{E}_{k+1})=\min_{\alpha}(\mathbb{E}_{k}) - \alpha_n(v-\mathbf{U}^{T}\alpha_k).
    \label{fun_energy_prelsm_next}
\end{equation}
Applying Cholesky decomposition to $\mathbf{G}_k$ gives $\mathbf{G}_k=\mathbf{R}_k^T\mathbf{R}_k$, where $\mathbf{R}_k$ is the upper triangular matrix. The factor for the next layer is:
\begin{equation}
    \mathbf{R}_{k+1} = \begin{bmatrix}
        \mathbf{R}_{k} & {\mathbf{R}_{k}^{-1}}^T\mathbf{U}\\
        0 & \sqrt{s}\\
    \end{bmatrix}.
    \label{fun_state_prelsm_next}
\end{equation}
Triangular solves are used instead of explicitly forming $\mathbf{G}_k^{-1}$. This reduces the coefficient solving cost of a layer extension from $O(m^3)$ to $O(m^2)$, giving $O(whM^3)$ prefix-search work apart from the first layer. The first layer is scored by dot products and projection. Subsequent states use ${\mathbf{C}_1^r}^{T}\mathbf{C}_s^r$ and ${\mathbf{C}_s^r}^{T}\mathbf{C}_s^r$, avoiding a large ${\mathbf{C}_1^r}^{T}\mathbf{C}_1^r$ table. Here $s$ in $\mathbf{C}_s^r$ denotes the other smaller layers.

\subsubsection{Local Refinement of Codes and Coefficients}
The beam search result are further refined by the following methods: Iterated local search (ILS) perturbs the code assignments to escape the current local optimum, iterated conditional modes (ICM) subsequently refines a layer while holding the others fixed, and LSM refits the coefficients for the updated discrete codes. The first-layer code remains fixed to preserve IVF membership. 

The best code index $j$ in layer $m$ is selected by cosine similarity $\mathbb{S}$ between each root codeword $\mathbf{c}^r$ and the current residual $\mathbf{r}_m^r$ of layer $m$, excluding the contributions of other layers. Omitting the shared residual reciprocal, we have:
\begin{align}
    &\arg\max_{\mathbf{c}_{m,j}^r}(\mathbb{S}) = \arg\max_{\mathbf{c}_{m,j}^r}\frac{({\mathbf{r}_m^r})^T\mathbf{c}_{m,j}^r}{\|\mathbf{c}_{m,j}^r\|}, \nonumber\\
    &r_m^r = \mathbf{x} - \alpha_1\mathbf{c}_1^r - \sum_{t=2,t\ne m}^{M}\alpha_t\mathbf{c}_t^r.
    \label{fun_batch_icm_condition}
\end{align}

After a code change, LSM recomputes $\alpha$, and the resulting energy is compared with the original one. Only the perturbations that successfully decrease the energy will be accepted. To speed up evaluating MSE after ILS, ICM, and LSM, we decouple the first layer to avoid calculating ${\mathbf{C}_1^r}^T\mathbf{C}_1^r$:
\begin{align}
    \mathbb{E} 
    &= \|\mathbf{x}\|^2 -2\mathbf{x}^{T}(\alpha_1\mathbf{c}_1^r+\sum_{m=2}^{M}\alpha_m\mathbf{c}_m^r) +\alpha_1^2\|\mathbf{c}_1^r\|^2 \nonumber\\
    &+ 2\alpha_1{\mathbf{c}_1^r}^T\sum_{m=2}^{M}\alpha_m\mathbf{c}_m^r + \sum_{m=2}^{M}\sum_{t=2}^{M}\alpha_m\alpha_t{\mathbf{c}_m^r}^T\mathbf{c}_t^r.
    \label{fun_icm_energy_decouple}
\end{align}

If an ICM round fails to improve a code, LSM is skipped to save computational cost. An early-stopping check on ICM also reduces overhead. The worst-case complexity remains $O(M\cdot M^3)$, although this case is never reached in practice. In fact, the batched basic encoding is not the bottleneck, while linked encoding is much slower because of the amplification factor of the neighbor number.

\subsubsection{Root Codebook Update}
With assignments and coefficients fixed, the root codebook update is a weighted dimension-wise LSM. Let $H^r=h_1^r+(M-1)h$, we have:
\begin{equation}
  \min_{\mathbf{C}^r}\|\mathbf{X}-\mathbf{C}^r\mathbf{A}\|_F^2,
  \qquad \mathbf{A}\in\mathbb{R}^{H^r\times N},\mathbf{X}\in\mathbb{R}^{d\times N},
  \label{fun_root_codebook_objective}
\end{equation}
where $\mathbf{A}$ is the weighted matrix formed from $\mathbf{B}$ and $\alpha$. This is similar to LSQ++ \cite{martinez2018lsq++}, which accumulates sparse-matrix statistics without explicitly constructing sparse matrices. The difference is that STLQ uses a weighted matrix instead of a binary one.

For $\mathbf{C}^r$, the large $h_1$ makes the normal equations expensive, so we solve the small layers first with a Schur-complement approach, and then back-substitute for $\mathbf{C}_1^r$. Partition $\mathbf{C}^r=[\mathbf{C}_1^r,\mathbf{C}_s^r]$ into first-layer and small-layer blocks, let $\mathbf{U}=\mathbf{A}\mathbf{A}^T$ and $\mathbf{V}=\mathbf{X}\mathbf{A}^T$, we have:
\begin{align}
    &\mathbf{C}_s^r=
        (\mathbf{V}_s-\mathbf{V}_1\mathbf{U}_{11}^{-1}\mathbf{U}_{1s})
        (\mathbf{U}_{ss}-\mathbf{U}_{s1}\mathbf{U}_{11}^{-1}\mathbf{U}_{1s})^{-1} \nonumber\\
    &\mathbf{C}_1^r = (\mathbf{V}_1-\mathbf{C}_s^r\mathbf{U}_{s1})\mathbf{U}_{11}^{-1}.
  \label{fun_solve_root_small_layer}
\end{align}

\begin{figure}[!htb]
  \centering
  \includegraphics[width=0.4\textwidth]{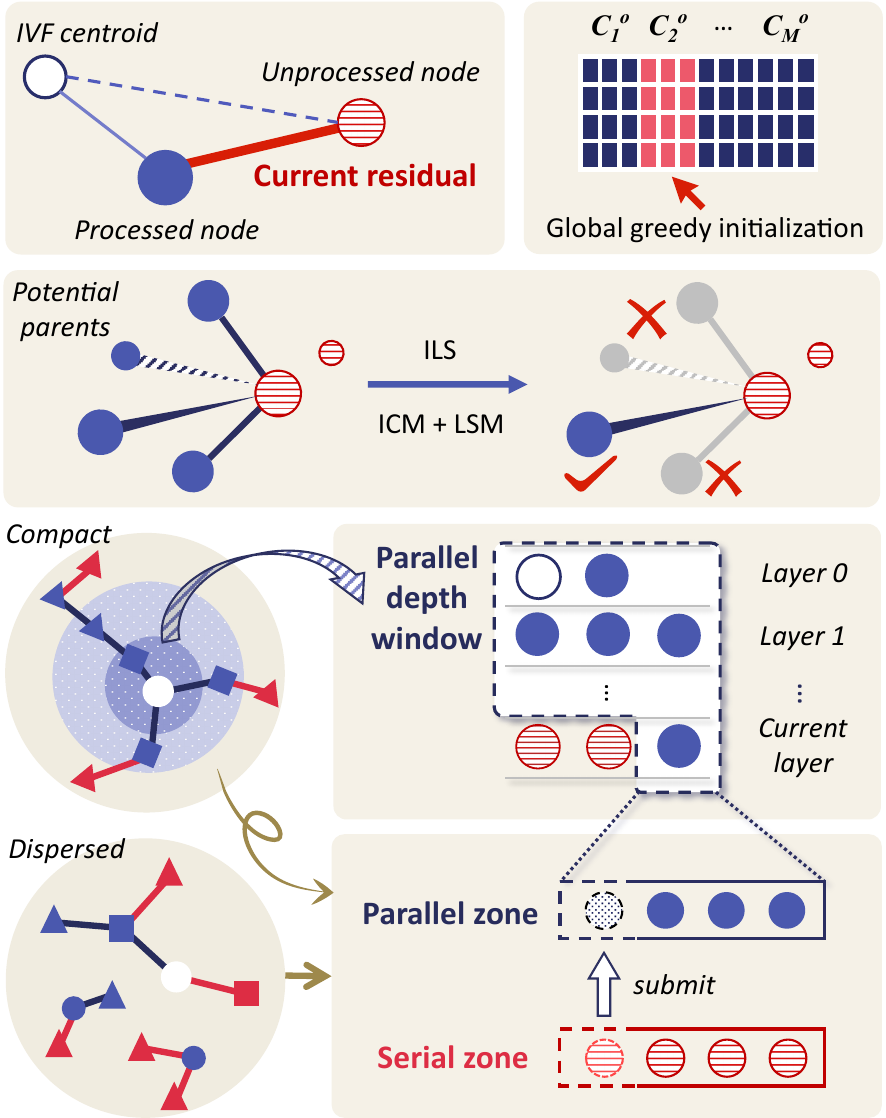}
    \caption{The diagram of linked encoding. The cluster is viewed from the IVF centroid. Its legend follows that in Fig. \ref{fig_technical_overview}. An unprocessed node evaluates feasible parent candidates, initializes and refines each relative residual, and accepts a link only if it decreases basic MSE. Already processed nodes form a parallel candidate pool to speed up with GPU, while the serial zone preserves parent-before-child dependencies.}
  \label{fig_linked_encoding_partial}
\end{figure}

\subsection{Linked Encoding: Exploiting Local References}
Linked encoding judges whether a feasible parent makes a vector easier to encode than its independent baseline. For each $\mathbf{x}_i$, STLQ first constructs a candidate parent set $\tilde{p}(i)$, then encodes the relative residual $\epsilon_{i|p}$ for each candidate using $\mathbf{C}^o$. The corresponding parent with the lowest MSE in candidate set is selected only if it is lower than the basic MSE. Fig. \ref{fig_linked_encoding_partial} illustrates candidate evaluation and the processing dependencies. The challenge is not only to find a low error linkage, but also to restrict the search so that the resulting representation is acyclic and affordable to construct.

\subsubsection{Feasible Parent Search under Dual Topologies}
Candidate parents must be encoded before their children, and their number must be limited because each candidate requires residual encoding. STLQ uses two topology strategies according to cluster geometry. In addition, random vector processing order is used rather than the greedy basic-MSE ascending order, which performed slightly worse in our experiments.

\paragraph{Outward-expanding topology}
For a compact cluster, vectors are partitioned into $\mathcal{L}$ nested hyper-conical layers around the IVF centroid direction according to their angular distribution. The innermost vectors are set as initial roots, after which processing proceeds from the inner layers to the outer layers. A vector may use parents from inner layers within the depth window $z$, or already processed vectors in its current layer. Hierarchical Navigable Small World (HNSW) \cite{malkov2018efficient} provides at most $k$ neighbors per layer, so each vector has at most $(z+1)k$ candidate parents. Since each candidate parent is processed before the current vector, the generated topology is acyclic.

\paragraph{Virtual-Relay Topology}
For a dispersed cluster, vectors near the IVF centroid may not provide sufficiently accurate references. STLQ therefore generates a small ratio $\gamma_{v}$ of virtual relay nodes from a fuzzy neighborhood graph generated by Uniform Manifold Approximation and Projection (UMAP) \cite{mcinnes2018umap}, which are treated as root nodes. These virtual nodes represent local high-density regions. They are encoded with the first layer fixed as auxiliary root references, and are not returned as database results. Each real node may choose up to $k$ virtual nodes and $k$ already processed real nodes as candidate parents, giving at most $2k$ candidates.

\subsubsection{Residual Codebook Initialization}
Learning $\mathbf{C}^o$ requires linked residuals, but linked encoding itself needs a codebook. STLQ resolves this initialization dependency with a preliminary outward-expanding topology constructed using $\mathbf{C}^r$, where linked nodes are collected for solving $\mathbf{C}^o$ via LSM. This initialization is performed before the alternating training loop. Legacy, Fast, and Hybrid initialization modes trade offline cost against initialization quality, whose differences are described in Appendix. We only use Legacy mode in experiments.

\subsubsection{Greedy Initialization for Candidate Linkage}
Since each vector may evaluate multiple candidate parents, beam search would be expensive. STLQ instead initializes each relative residual $\mathbf{r}$ by a global greedy search across the unassigned layers of $\mathbf{C}^o$. Given the prefix of $k$ codes, it selects a new codeword $\mathbf{c}_n^o$ each time from the unassigned layers in $\mathbf{C}^o$, and computes the new projection coefficients $\alpha_n$, at a cost of $O(dMh+M^2h)$. Minimize $\mathbb{E}$ by maximizing the absolute $|\mathbb{S}|$:
\begin{align}
    \arg\min_{\mathbf{c}^o,\alpha_n}(\mathbb{E}_{k+1})
    &=\arg\min_{\mathbf{c}^o,\alpha_n}\|\mathbf{r}_k - \alpha_n\mathbf{c}_n^o\|^2 \nonumber\\
    &= \arg\max_{\mathbf{c}^o,\alpha_n} \left|\frac{{\mathbf{r}_k}^T\mathbf{c}_n^o}{\|\mathbf{c}_n^o\| }\right|, \quad \alpha_n = \frac{{\mathbf{r}_k}^T\mathbf{c}_n^o}{\|\mathbf{c}_n^o\|^2}.
\end{align}
After all codewords have been selected, the scaling coefficients are updated by LSM. Unlike basic encoding, all code assignments including the first layer can be modified, because the linked representation does not undertake IVF routing. Negative coefficients are also allowed through absolute $|\mathbb{S}|$.

\subsubsection{Joint Refinement and Parent Selection}
\paragraph{Candidate Refinement}
For each candidate parent, ILS and ICM refine the greedy code, and LSM refits the coefficients after a code change. For a vector $\mathbf{x}_i$ and its candidate parent $p(i)$, ILS and ICM are similar to Eq. \ref{fun_batch_icm_condition} in basic encoding. The differences are that absolute cosine similarity $|\mathbb{S}|$ is used, and the parent reconstruction should follow Eq. \ref{fun_x_reconstruction}, which means root nodes should use $\mathbf{C}^r$. Several LUTs can be applied to speed up finding the best code $j$ in $m$-th layer, where $\epsilon_{i|p}$ is the relative residual:
\begin{align}
    \arg\max_{\mathbf{c}_{m,j}^o}(\mathbb{|S|})=\arg\max_{\mathbf{c}_{m,j}^o}\left|\frac{({\mathbf{r}_m^o})^T\mathbf{c}_{m,j}^o}{\|\mathbf{c}_{m,j}\|}\right| ,\nonumber\\
    \mathbf{r}_m^o = \epsilon_{i|p} - \sum_{t=1,t\neq m}^{M}\alpha_t\mathbf{c}_t^o.
    \label{fun_linked_icm}
\end{align}

\paragraph{Parent Selection and Acceptance}
Since $\mathbf{C}^o$ has a small $h$, the MSE between $\epsilon_{i|p}$ and the refined result is simple to get with the LUT:
\begin{align}
    &\min_{\mathbf{c}^o}(\mathbb{E}^o_{i|p})= \min_{\mathbf{c}^o}\|\epsilon_{i|p}\|^2 - 2\epsilon_{i|p}^T\sum_{m=1}^{M}\alpha_m\mathbf{c}_m^o \nonumber\\
    &+ \sum_{m,t=1}^{M}\alpha_m\alpha_t{\mathbf{c}_m^o}^T\mathbf{c}_t^o.
\end{align}
Let $\mathbb{E}^o_{i|p^*}$ be the lowest linkage MSE of candidate parent $p^*(i)$, $\mathbb{E}_i^r$ be the basic MSE. A linkage is accepted only if
\begin{equation}
\mathbb{E}^o_{i|p^*} < \mathbb{E}_i^r.
\end{equation}
Otherwise, $\mathbf{x}_i$ remains a root node and retains its basic encoding.

\subsubsection{Residual Codebook Update}
With topology, codes, coefficients, and $\mathbf{C}^r$ fixed, the residual codebooks $\mathbf{C}^o$ are updated by weighted LSM on the linked nodes:
\begin{align}
    &\min_{\mathbf{c}^o}\sum_{D_i\ge 1} \lambda_i\|\epsilon_{i|p} - \sum_{m=1}^{M}\alpha_{i,m}\mathbf{c}_{i,m}^o\|^2,\nonumber\\
    & \lambda_i =
            \begin{cases}
            1, & \mathbf{x}_i \in \text{compact cluster}\\
            \tau\in(0,1]. & \mathbf{x}_i \in \text{dispersed cluster}
            \end{cases}
    \label{fun_residaul_codebook_update}
\end{align} 

\subsubsection{Parallel Implementation on GPU}
Basic encoding is processed in batches, while linked encoding contains sequential dependencies, because a parent must be encoded before its child. STLQ therefore aggregates already processed nodes. Fig. \ref{fig_linked_encoding_partial} illustrates this concept. For the outward-expanding topology, nodes in completed inner layers can be evaluated as the potential parents in parallel by GPU, while the current processing layer is divided into a parallel zone and a serial zone. Once a serial node finishes evaluating all parent candidates, it moves to the parallel zone. The virtual-relay topology is handled similarly. Since the topology is processed cluster-wise, it is also suitable for multi-GPU architectures.

\subsection{Alternating Optimization and Database Encoding}
After root and residual codebooks initialization, each training iteration rotates the training data as $\mathbf{X}'=\mathbf{R}_{t-1}\mathbf{X}$. Basic encoding then refreshes the independent representations and updates $\mathbf{C}^r$. Linked encoding constructs the topology and updates $\mathbf{C}^o$. The resulting reconstructions are used in the OPP update of $\mathbf{R}_t$.

After training, the learned codebooks and rotation are fixed, and the base set is encoded through one round of basic and linked phases. The resulting coefficients and forest are then compressed as the description in Section \ref{lab_bit_control}. Note that the basic encoding phase generally follows the same procedure as training, but with a different selection rule. For training, only codewords with positive $\mathbb{S}$ are used to preserve the learned codeword directions. In contrast, for base set encoding, selection is based on absolute similarity $|\mathbb{S}|$, which enhances expressiveness through the use of a negative $\alpha$.

\subsection{Bit Budget Control}
\label{lab_bit_control}
The linked representation reduces distortion at the cost of additional coefficients and topology metadata. To satisfy the storage budget, compression algorithms are required to control the bit budget.

\subsubsection{Coefficient Compression}
To compress the float $\alpha$ as a quantized $\hat{\alpha}$, a lossy compression is applied for each layer in each cluster, so there would be $M h_1^r$ scaling factors saved for rescaling $\hat{\alpha}$. After truncating a fraction of outliers, float $\alpha$ are linearly mapped to a 6-bit integer range, and a lightweight ICM then refines the values within a small local candidate set to reduce loss. This gives about 6 bits per $\hat{\alpha}$ with slight accuracy loss. Because the coefficients have skewed distributions, Huffman coding is further applied to enable $5$ bits per $\hat{\alpha}$. At the retrieval stage, a base LUT is computed first, and then per-layer scale factors are multiplied back for each queried cluster.

\subsubsection{Topology Compression}
For a directed-forest containing $n$ nodes in a cluster, the Level-Order Unary Degree Sequence (LOUDS) \cite{jacobson1989space} algorithm compresses them into a bitvector with $2n+1$ bits. Therefore, it reduces an integer parent index in $\mathcal{P}$ to nearly 2 bits. On retrieval, this bitvector can be decoded in sequence without losing too much QPS, or be deserialized into $\mathcal{P}$ in advance to trade memory for higher QPS.

\subsubsection{Bit Accounting}
We set $M=5$ and $M=10$ to roughly correspond to conventional 64 and 128 bits. For example, when $M=5$, STLQ uses five bytes of codes and five coefficients of around five bits each, resulting in roughly 65 payload bits. In practice, the result depends on the linked ratio $\gamma_l$ and the virtual ratio $\gamma_v$. We therefore report the payload bits $\beta_p$ per vector without structural overhead, and the total bits $\beta_t$ per vector considering more information.

The payload bits consist of root codes, linked codes, and coefficients, which are listed in order as follows:
\begin{equation}
   \beta_p = 8(1-\gamma_l)(M-1) + 8M\gamma_l + M\hat{\alpha}_b^t, 
\end{equation}
where $\hat{\alpha}_b^t$ denotes the average bit usage of an integer coefficient in Huffman encoding, including the Huffman data, header, and linear scale factors. These $\hat{\alpha}_b^t$ contents are amortized by total nodes, including virtual nodes and real nodes. Because the first layer of root nodes is structural IVF, it is not accounted into $\beta_p$ by $(M-1)$.

The $\beta_t$ additionally includes the IVF, LOUDS, virtual codes, and one byte database norm index for $\| \hat{\mathbf{x}} \|^2$ in ADC, which are listed in order in the first line of Eq. \ref{fun_total_bit_budget}:
\begin{align}
   \beta_t &= \log_2(h_1^r) + \frac{1}{N}\sum_{i=1}^{h_1^r}(2n_i+1) + 8\gamma_v(M-1) + 8 \nonumber\\
   &+8(1-\gamma_l)(M-1) + 8M\gamma_l + M\hat{\alpha}_b^r, 
   \label{fun_total_bit_budget}
\end{align}
where $n_i$ is the number of all nodes including virtual ones in each cluster. The coefficient bits $\hat{\alpha}_b^r$ are amortized by real nodes, and include the contribution of virtual nodes.

\subsection{Linked-ADC Function}
The query retrieval is based on the linked-ADC function in Eq. \ref{fun_adc_score}, where the ${\mathbf{q}'}^T\hat{\mathbf{x}}$ term can be calculated along the level order parent array $\mathcal{P}$ with $O(n)$ cost by the LUTs between $\mathbf{q}'$ and $\mathbf{C}^r,\mathbf{C}^o$. Compared with normal additive-ADC, the extra constant factor is $M+1$ for $M$ coefficient multiplications, and one parent accumulation. Although the $\| \hat{\mathbf{x}} \|^2$ term can be trivially pre-computed, using LUTs can speed up this offline step for large-scale databases. We provide this detail in Appendix. After obtaining $\| \hat{\mathbf{x}} \|^2$, using k-means algorithm with one byte index to quantize them.

\subsection{Discussion}
\label{lab_discussion}
\subsubsection{Linkage Acceptance Criterion}
In linked encoding, we require a strict MSE reduction before accepting a linkage. We also tested a slightly relaxed criterion with a mild regularizer to encourage more linked relationships. With the smaller setting $M=5$, the relaxed rule indeed gives a slight improvement on recall, suggesting better local resolution, although MSE becomes slightly worse as expected. However, $M=10$ shows the opposite trend with slightly lower recall. We infer that the local resolution is already sufficient for $M=10$, and the extra error starts to dominate. Consequently, we just keep the strict acceptance rule.

\subsubsection{Two-Phase Encoding Balance}
The beam width $w$ in basic encoding is typically small in STLQ, and is often set as $w=2$. A larger $w$ improves the basic encoding, but reduces the linked ratio in linked encoding, which limits the gain from local resolution and ultimately impacts recall. So the choice of $w$ should account for the overall interaction, rather than optimizing only the basic phase.

This also explains why we do not use the external IVF residual mode. STLQ lets the first layer of the root codebook undertake the IVF routing function directly in an internally IVF manner. By contrast, CompQ \cite{ozan2016competitive}, QINCo \cite{huijben2024residual}, and the Faiss library\cite{douze2024faiss} use the conventional external IVF. While IVF-based residual structure helps basic encoding, it encounters the same problem as that in beam-width.

\subsubsection{Support for Dynamic Database Updates} 
Although STLQ uses a differential encoding mechanism, its cluster-wise encoding model has limit influence on dynamic updates. The corresponding coefficients and LOUDS topology are updated cluster-wise. For insertion, a lightweight strategy encodes only new vectors using existing or previously inserted nodes as candidate parents, while a full cluster re-encoding can be used for better global optimization. For deletion and update, a leaf node can be operated directly, whereas an internal parent requires re-encoding its affected descendants.

\subsection{Complexity Analysis}
\label{lab_complexity_analysis}

Table \ref{tab_complexity} summarizes the operation complexities in STLQ. See Table \ref{tab_symbol} for symbol definitions. Define $H_s^r=(M-1)h$, $H^r=h_1^r+H_s^r$, and $H^o=Mh$. Symbol $n$ represents the cluster-wise complexity. Symbols $T_{sk},T_{ils},T_{icm},T_{\alpha}$ denote the iterations for spherical k-means, ILS, ICM, and light ICM in coefficient compression, respectively.

\subsubsection{Basic Encoding Complexity}

Root codebook initialization is performed once before iterative training. For root ILS/ICM, preparing the $m$-th layer residuals in turn takes $O(hM^2)$. The $O(M^4)$ term represents the worst case that is hardly reached, where all $M$ updates are accepted and each requires an $O(M^3)$ LSM cost.
\subsubsection{Linked Encoding Complexity}
Let $\gamma_d$, $\gamma_v$, and $\gamma_l$ denote the ratios of dispersed clusters, virtual nodes, and accepted linked nodes, respectively. There are $N\gamma_v$ virtual nodes in total, and $n_v=n\gamma_v/\gamma_d$ virtual nodes per dispersed cluster on average. The $\gamma_l$ is statistically obtained from the quantization result. The ILS/ICM complexities in Table \ref{tab_complexity} are given per vector. For basic encoding, the amplification factor is $N$. For linked encoding, HNSW provides $k^c \le \min\{n-1,(z+1)k\}$ and $k^d \le \min\{n+n_v-1,2k\}$ neighbors for each node in a compact cluster and a dispersed cluster, respectively. Thus, global greedy initialization and linked ILS/ICM are performed up to $nk^c$ and $nk^d$ times per cluster. For residual codebook initialization, the legacy mode replaces the $H^o$ term in global greedy initialization and ILS/ICM with $H^r$. The fast mode just uses $H^o$. Therefore, the legacy mode is more expensive when $h_1^r$ is large for large-scale data.

\begin{table*}[t]
\centering
\caption{Complexity of the main operations in STLQ, where some constant factors are retained for clarity}
\label{tab_complexity}
\begin{tabularx}{\textwidth}{lllX}
\toprule
\textbf{Operation} & \textbf{Complexity} & \textbf{Operation} & \textbf{Complexity} \\
\midrule

\textbf{Basic Encoding Section} & & Root codebook initialization & $O(T_{sk}dNH^r)$  \\
Coefficient LSM & $O(NM^3)$ & Beam search with prefix-LSM & $O(N(h_1^r + w hM^3))$  \\

Root ILS/ICM per node & $O(T_{ils}T_{icm} (H_s^r M+M^4))$& Root codebook update & $O(N(dM+M^2) + (H_s^r)^3+(H_s^r)^2 d + h_1^r(H_s^r+d))$ \\

\midrule
\textbf{Linked Encoding Section}& & HNSW build and search & $O(ndk\log n)$ \\

Global greedy initialization & $O(H^o(d+M))$ & Virtual nodes augmentation & $O(nk^2+ (n \gamma_v/\gamma_d) dk + (n \gamma_v/\gamma_d) H_s^r(d+M))$ \\

Linked ILS/ICM per parent & $O( T_{ils}T_{icm}(H^oM+M^4) )$ & Residual codebook update & $ O(N\gamma_l(dM+M^2) +(H^o)^3+(H^o)^2d)$ \\

Orthogonal rotation & $O(Nd^2+d^3)$ & Reconstruct $\mathbf{x}$ for rotation & $O(NdM)$\\

\midrule
\textbf{Compression Section} && Coefficient quantization & $O(M(n \log n + n T_{\alpha}))$ \\

LOUDS serialization & $O(n)$ & Huffman coding & $O(nM)$\\

\midrule
\multicolumn{4}{l}{\textbf{Retrieval Section}} \\

Norm pre-computation & $O(NM^2D_{\max})$& LUTs of $\mathbf{C}^r$ and $\mathbf{C}^o$ & $O((H_s^r)^2+(H^o)^2+H_s^rH^o)$ \\

Linked-ADC scan & $O(n)$ & LUTs of $\mathbf{q}'$ and $\mathbf{C}^r,\mathbf{C}^o$ & $O(d(H^r+H^o))$ \\

\bottomrule
\end{tabularx}
\end{table*}

\section{Experiments}
\label{lab_experiments}

\subsection{Experiment Settings}
\subsubsection{Baselines and datasets}
\label{lab_baselines_and_datasets}
We compare STLQ with the shallow MCQ methods OPQ \cite{ge2013optimized} and LSQ++ \cite{martinez2018lsq++}, the neural MCQ methods DR-UNQ \cite{noh2023disentangled} and QINCo \cite{huijben2024residual}, and the data-oblivious method RaBitQ \cite{gao2024rabitq}. OPQ and LSQ++ are implemented with Faiss \cite{douze2024faiss} using residual IVF coarse quantizer, which is also adopted by DR-UNQ and QINCo. In contrast, STLQ integrates the IVF coarse quantizer into its root codebook, as discussed in
Section \ref{lab_discussion}. We additionally profile a GPU implementation of LSQ++. Since DR-UNQ is not open-source, we report the results from its paper. For QINCo, we evaluate the released pretrained model and IVF index on BigANN. QINCo2 is omitted because no corresponding pretrained database is available, and reproducing it requires multiple GPUs, which is beyond our resources. All the cited data are marked with underline.

We evaluate performance on BigANN (also known as SIFT1B), \cite{jegou2011searching}, SIFT1M \cite{jegou2010product}, GIST1M \cite{jegou2010product}, and DEEP1M \cite{babenko2016efficient} datasets. The features are listed in Table \ref{tab_datasets}. We only use a subset from the 100M train set of BigANN for training. SIFT1M is an independent small dataset, rather than the 1M data from BigANN. DEEP1M uses the first 1M base set and the first 500k train set from DEEP1B.

Because the compared baselines usually obtain better recall, MSE, and QPS with larger IVF centroids, we set the IVF numbers to their preferred values. In contrast, STLQ needs enough nodes per cluster to capture the topological relationship, so a smaller IVF number is preferred. For BigANN, DR-UNQ uses 100M train set and $2^{12}\times2^{12}$ IVF centroids, while other baselines use 10M train set and 1M IVF centroids. STLQ uses $h_1^r=65\text{k}$ or $h_1^r=100\text{k}$, and 10M to 50M train set for sensitivity analysis. For 1M datasets, QINCo uses no IVF, other baselines use 1024 IVF centroids as default, while STLQ uses $h_1^r=256$.

\subsubsection{Implementation and Metrics}
Experiments were run on a Linux workstation with dual-CPU (each Intel(R) Xeon(R) Gold 6226R CPU @ 2.90GHz has 16 cores and 32 threads) and 512 GB of RAM (64 GB per DDR4 DIMM at 2934 MT/s). An NVIDIA 4090 GPU is used in offline stages.

Evaluation metrics include the top-1 ground truth Recall@k (R@k), QPS, MSE, payload bits $\beta_p$, total bits $\beta_t$, and disk size. The nominal 8 bytes (8B) and 16 bytes (16B) of $\beta_p$ are shorthand for $M=5$ and $M=10$, respectively.

\subsubsection{Parameter Settings of STLQ}

\paragraph{Topology search parameters} The topology search strength is controlled by three parameters: the number of topology layers $\mathcal{L}$, the depth-window size $z$, and the HNSW neighbor number $k$. For BigANN training, the small (S), medium (M), and large (L) settings of $(\mathcal{L},z,k)$ are $(10,6,15)$, $(12,10,10)$, and $(12,11,15)$, respectively. For base-set encoding, the small (S) and large (L) settings are $(12,8,20)$ and $(16,15,20)$. For example, the \texttt{SL} suffix denotes small training strength and large base set strength.

\paragraph{Dataset-dependent parameters} The number of IVF clusters, train set size, number of training iterations, and residual codebook initialization mode are selected according to the dataset scale. We use $T=25$ training iterations by default, except for GIST1M, which uses $T=15$ because it is more sensitive to overfitting. For example, \texttt{IVF65k-30M-LL} for BigANN denotes 65K IVF clusters, 30M training vectors, and the large topology search setting for both training and base set encoding. For million-scale datasets, the topology search settings are selected separately for each dataset.

\paragraph{Fixed optimization parameters} We fix the virtual ratio to $\gamma_v=0.05$, the beam width to $w=2$, and the maximum linkage depth to $D_{max}=40$. Following LSQ \cite{martinez2016revisiting}, for basic encoding, we set $T_{icm}=4$ and $T_{ils}=8$ for training, while setting $T_{icm}=4$ and $T_{ils}=32$ for base set encoding. For linked encoding, we set $T_{icm}=1$ and $T_{ils}=4$ for both training and base set encoding. Each ILS round perturbs three code indices. Set the dispersed cluster ratio to $\gamma_d=0.4$, and the residual codebook update weight to $\tau=0.3$.

\begin{table}[!htb]
    \centering
    \caption{Features of datasets}
    \label{tab_datasets}
    \begin{tabular}{lllll}
        \toprule
        Dataset & Train set    & Base set         & Query set & Dimension \\
        \midrule
        SIFT1M  & $10^5$    & $10^6$      & $10^4$  & $128$\\
        BigANN  & $1/3/5\times10^7$ & $10^9$  & $10^4$  & $128$\\
        GIST1M  & $5\times10^5$    & $10^6$      & $10^3$   & $960$\\
        DEEP1M & $5\times10^5$  & $10^6$     & $10^4$  & $96$\\
        \bottomrule
    \end{tabular}
\end{table}

\subsection{Quantization Quality and Offline Cost}
\subsubsection{Reconstruction Quality}
\begin{figure}[!htb]
  \centering
  \begin{minipage}[b]{1.0\linewidth}
  \subfloat[]{\includegraphics[width=0.48\linewidth]{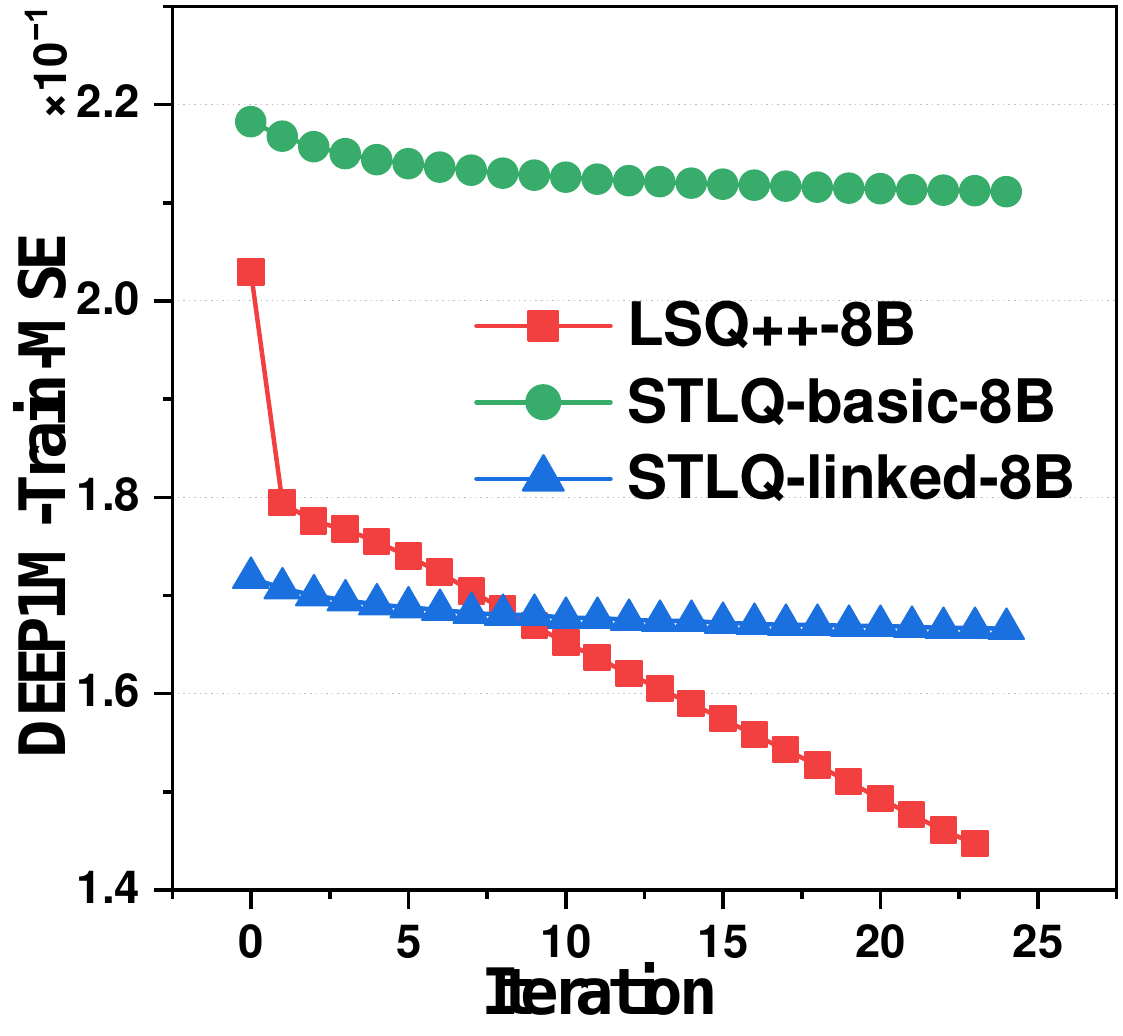}}
  \subfloat[]{\includegraphics[width=0.49\linewidth]{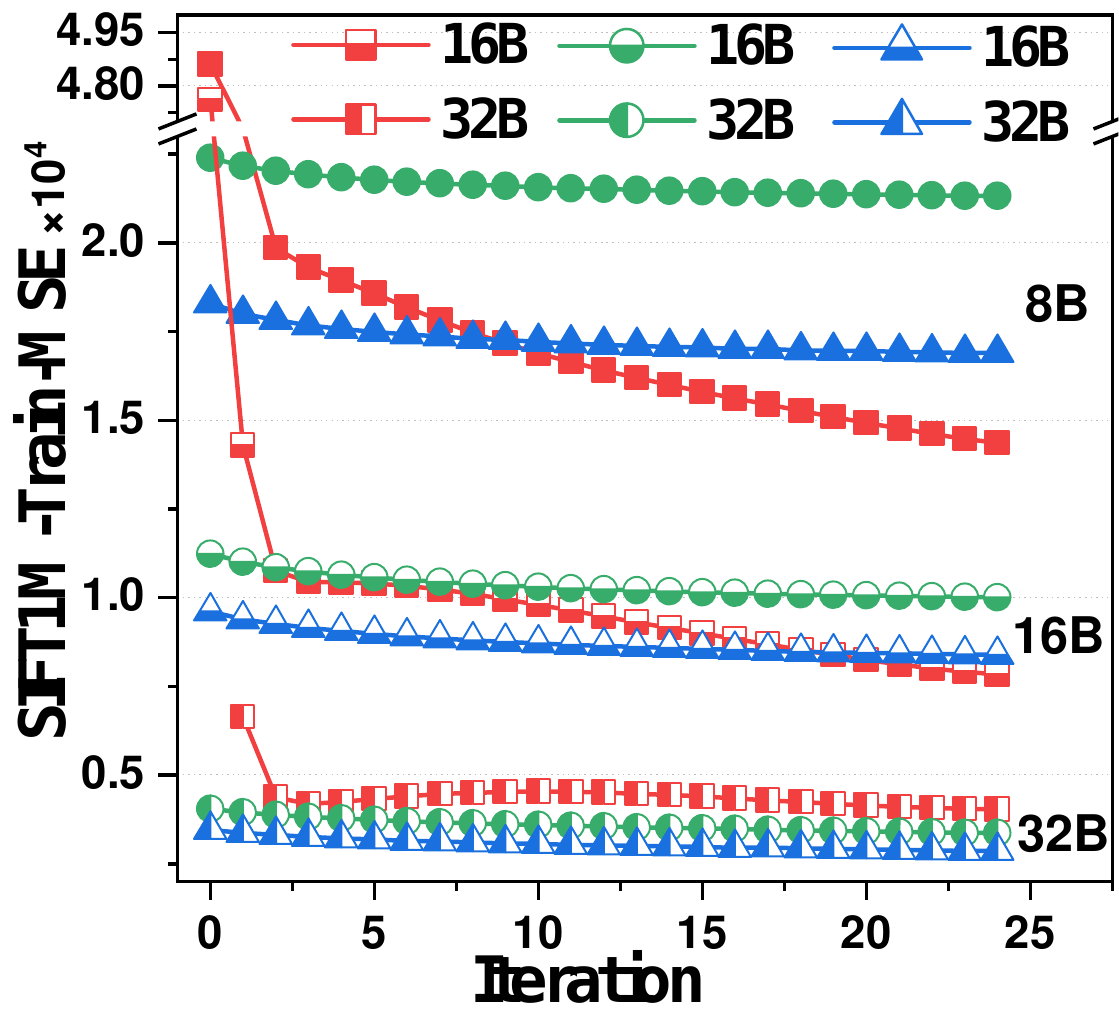}}
  \end{minipage}
  \caption{Train set MSE on two encoding phases of STLQ (circle markers represent basic encoding, and triangle markers represent linked encoding) and LSQ++ (square marker) with iterations on (a) DEEP1M and (b) SIFT1M.}
  \label{fig_mse_comparation}
\end{figure}

\paragraph{Training Convergence} As STLQ uses an LSQ-style encoding backbone, we compare the training convergence with LSQ++. Fig. \ref{fig_mse_comparation} reports the train set MSE of basic and linked encoding. Linked encoding further reduces the MSE of basic encoding, showing that the topological representation captures local structures. Although the training MSE of STLQ decreases slowly, stopping too early is not beneficial for recall. For SIFT1M and DEEP1M, at least 20 iterations with stronger parameters usually produce better recall.

\paragraph{MSE of Base Set} Table \ref{tab_mse_all_compare} gives the base set MSE report. Although the MSE of STLQ on one million-scale datasets is higher than LSQ++ at 8B quantization, it still produces a larger recall. Compared to LSQ++, the MSE level at 16B quantization is usually reversed, which is likely because the higher bitrate better captures the fine-grained local structure. On BigANN, STLQ exceeds other shallow methods, which indicates that the denser nodes within a cluster are more conducive to a good linked graph.

\begin{table}[!htb]
    \centering
    \caption{The base set MSE of MCQ methods. QINCo uses its L16 version. STLQ uses IVF65k-30M-LL for BigANN 8B, IVF100k-30M-SL for 16B}
    \label{tab_mse_all_compare}
    \begin{threeparttable}
    \begin{tabular}{lcccccc}
        \toprule
        & \multicolumn{2}{c}{\bf BigANN/$\cdot 10^4$}   & \multicolumn{2}{c}{\bf DEEP1M/$\cdot 10^{-2}$} &\multicolumn{2}{c}{\bf GIST1M/$\cdot 10^{-1}$} \\
        \cmidrule(lr){2-3} \cmidrule(lr){4-5} \cmidrule(lr){6-7}
        & 8B &16B & 8B &16B & 8B &16B\\
        \midrule
        OPQ   & 1.918 & 1.106  & 19.54 & 9.589 & 6.629  & 5.059\\
        LSQ++ & 1.521 & 0.8626 & 16.00 & 8.740 & 5.878  & 4.549 \\
        QINCo & * & * & \underline{12.00} & \underline{5.000}  & /  & /\\
        STLQ  & 1.313 & 0.6662 & 16.17 & 7.997 & 5.925  & 4.515\\
        \bottomrule
    \end{tabular}
    \begin{tablenotes}
      \item[*] No QINCo MSE is available for BigANN. The starred values are \underline{1.120} at 8B and \underline{0.3200} at 16B from BigANN1M as references
    \end{tablenotes}
\end{threeparttable}
\end{table}

\subsubsection{Storage Cost}
Table \ref{tab_storage_compare} summarizes the payload bit budget $\beta_p$, total bit budget $\beta_t$, and the disk size on BigANN-16B. Global or auxiliary structures such as HNSW index, codebooks, and neural network model are not included in $\beta_t$ but are included in the disk size. The disk size is counted after removing the duplicated and temporary artifacts. Since OPQ does not need extra database norms in Cartesian product codebook structure, it has the smallest disk usage. For STLQ, the decompositions of $\beta_p$ and $\beta_t$ are listed in Table \ref{tab_storage_split_STLQ}. Other MCQs use a fixed $\beta_p=128$ bits, and do not have amortized LOUDS, coefficients, and virtual bit contributions. The $\beta_p$ and $\beta_t$ of STLQ are influenced by the virtual ratio, linked ratio, and the coefficient compression. For example, STLQ only needs $\beta_p=125.65$ bits on GIST1M-16B because of more compressible coefficients.

\begin{table}[!htb]
    \centering
    \caption{Bit budgets and disk size on BigANN, using 16B of $\beta_p$. All methods use HNSW-based IVF except for DR-UNQ. STLQ uses IVF65k-50M-SS. See Section \ref{lab_baselines_and_datasets} for more settings }
    \label{tab_storage_compare}
    \begin{tabular}{lccccc}
        \toprule
        & OPQ & LSQ++ & QINCo & STLQ & DR-UNQ\\
        \midrule
        $\beta_p$ /bit & 128.00 & 128.00  & 128.00  & 129.70 & 128.00 \\
        $\beta_t$ /bit & 147.93 & 155.93  & 155.93  & 160.30  & 160.00 \\
        Disk /MB    & 19829.1 & 20768.7  & 20833.4  & 21153.8 & /  \\
        \bottomrule
    \end{tabular}
\end{table}

\begin{table}[!htb]
    \centering
    \caption{The decomposition of $\beta_p$ and $\beta_t$ in STLQ. $\beta_t$ omits the same code in $\beta_p$, IVF number, and 8 bits index of norm}
    \label{tab_storage_split_STLQ}
    \begin{tabular}{lcccccc}
        \toprule
        &\multicolumn{3}{c}{$\beta_p$ /bit} &\multicolumn{3}{c}{Part of $\beta_t$ /bit}\\
        \cmidrule(lr){2-4} \cmidrule(lr){5-7}
        &$M\hat{\alpha}_b^t$ & \makecell{root\\code} & \makecell{linked\\code} & $M\hat{\alpha}_b^r$ & \makecell{\small virtual\\code} & \makecell{LO-\\UDS}\\
        \midrule
        BigANN-16B & 50.02 & 2.945 & 76.73  & 51.88  & 2.672  & 2.074 \\
        GIST1M-16B & 47.68 & 18.19 & 59.78  & 49.67  & 3.015  & 2.084 \\
        DEEP1M-8B  & 25.13 & 0.9728 & 38.79  & 25.97  & 1.074  & 2.067 \\
        SIFT1M-8B  & 25.39 & 0.3456 & 39.57  & 26.67  & 1.623  & 2.102 \\
        \bottomrule
    \end{tabular}
\end{table}

\subsubsection{Offline Training and Encoding Cost}
Table \ref{tab_train_time_compare} reports the offline building time on BigANN dataset. Due to implementation differences, the data are for reference only. OPQ uses 50 training iterations, where GPU is used only for clustering IVF. LSQ++ uses 25 training iterations. STLQ uses IVF100k-30M-SL. The underlined time of the QINCo L16 model is taken from QINCo2 \cite{vallaeys2025qinco2}. STLQ has higher offline cost than other shallow MCQs, mainly because each vector evaluates multiple candidate parents during linked encoding. It can be alleviated by decreasing the topology search strength, using Fast initialization mode, and multiple GPUs.

\begin{table}[!htb]
    \centering
    \caption{End-to-end training and base set encoding time in hours (h) on BigANN. STLQ uses IVF100k-30M-SL and Legacy initialization mode. The QINCo data are cited from \cite{vallaeys2025qinco2}, using no IVF, and no base set encoding time is available}
    \label{tab_train_time_compare}
    \begin{tabular}{lccccc}
        \toprule
        &\makecell{Train\\/h}&\makecell{Base\\/h} &\makecell{Train\\size} & \makecell{IVF\\size} & \makecell{Known\\Hardware}\\
        \midrule
        OPQ-8B    &0.7 & 5.6 & 10M & 1M & \multirow{4}{*}{ \makecell{ one NVIDIA 4090\\GPU, Intel Xeon\\ Gold 6226R CPU }}\\
        OPQ-16B   &0.3 &  5.3 & 10M& 1M & \\
        LSQ++-16B &0.5 & 4.2 & 10M & 1M & \\
        STLQ-16B  &11.9 & 14.3 & 30M & 100k & \\
        \cmidrule(lr){6-6}
        QINCo-16B &\underline{284.9} & / & 10M & / & 8 Tesla V100 GPUs\\
        \bottomrule
    \end{tabular}
\end{table}

Table \ref{tab_detail_time} further gives the detailed time of STLQ on Legacy initialization mode. The statistics include the processing speed of basic encoding per 1M vectors, linked encoding time for train set and base set, and root codebook update time per iteration. The Fast mode costs about the same as one training iteration, whereas the Hybrid mode takes an intermediate time.
The GPU acceleration ratio over CPU on SIFT1M is 8x for basic encoding processed in batch. For linked encoding on base set, the ratio is 25x, whereas the ratio reduces to 7x on the smaller train set because of the finer task.

\begin{table}[!htb]
    \centering
    \caption{Time breakdown in seconds (s) of STLQ using Legacy initialization mode with GPU. The 1M datasets use IVF-256. BigANN-8B uses IVF65k-30M-LL, 16B uses IVF100k-30M-SL. Abbreviations: initialization (Init.); iteration (Iter.)}
    \label{tab_detail_time}
    \begin{tabular}{lccccc}
        \toprule
        & \multirow{3}{*}{\makecell{\textbf{Basic} /s\\per 1M\\ vectors}} &\multicolumn{3}{c}{\textbf{Linked} /s} &\multirow{3}{*}{\makecell{\textbf{Update} \\$\mathbf{C}^r$ /s \\ per Iter.}}\\
        \cmidrule(lr){3-5}
        &&\multicolumn{2}{c}{\textbf{Train set}}&\multirow{2}{*}{\makecell{\textbf{Base}\\ \textbf{set}}}&\\
        \cmidrule(lr){3-4}
        & &  Init. &1 Iter. & &\\
        \midrule
        SIFT1M-8B   & 2.3 & 3.8   & 2.6  & 17   & 0.31  \\
        DEEP1M-8B   & 2.2 & 15    & 8.1  & 17   &0.69  \\
        GIST1M-16B  & 13  & 52    & 28   & 158  &5.5\\
        BigANN-8B   & 2.5 & 8789  & 597  &34079 &15\\
        BigANN-16B  & 8.3 & 10760 & 978  &51326 &52\\
        \bottomrule
    \end{tabular}
\end{table}

\subsection{Retrieval Performance}
\subsubsection{Recall}
Tables \ref{tab_recall_1m_datasets} and \ref{tab_recall_bigann} report top-1 recall at $k$, under the nominal 8B and 16B payload bit budgets $\beta_p$. R@1 is the most sensitive to ranking errors, R@10 reflects shortlist quality, and R@100 is less discriminative because most methods approach saturation. The comparison uses a competitive configuration for each method, with the IVF number, train set size, and model configuration. The controlled comparisons under a common IVF configuration and total bit budget $\beta_t$ are also given, more results are further reported in Section \ref{lab_parameter_sensitivity}.

\paragraph{Comparison with Shallow MCQs.}
STLQ achieves the highest R@1 among the evaluated shallow MCQs. On BigANN, the improvement percentage points of R@1 are 7.63 at 8B and 13.74 at 16B, respectively. The improvements are also consistent at R@10. These results show that the linked representation in STLQ improves ranking quality over independently encoded shallow MCQs, and remains effective at the billion-scale. Apart from the data in Table \ref{tab_recall_bigann}, we also test the 100k IVF centroids for LSQ++ and OPQ. Increasing the IVF number from 100K to 1M improves the R@1 of LSQ++ from 17.15 to 18.58 at 8B, but reduces it from 34.46 to 33.00 at 16B. In contrast, OPQ benefits consistently from the finer IVF partition at both code lengths. The Faiss implementation of LSQ++ omits the double-round initialization and multi-peak fitting used by its original Julia implementation, which may contribute to the observed sensitivity.

\paragraph{Comparison with Deep MCQs}
Compared with the deep MCQ methods DR-UNQ and QINCo, STLQ is competitive but does not uniformly achieve the highest recall. On BigANN, QINCo obtains the highest R@1 and R@10, while STLQ narrows the recall gap between shallow and deep MCQs. Since maximum recall alone does not capture query efficiency, Section \ref{lab_recall_qps_tradeoff} further compares the complete recall-QPS tradeoff.

\begin{table*}[!htb]
    \centering
    \caption{The top-1 ground truth R@k on one million-scale datasets. See Section \ref{lab_baselines_and_datasets} for other baseline settings. Abbreviations: R@k ($\mathcal{R}$k); OPQ ($\mathcal{M}1$); LSQ++ ($\mathcal{M}2$); DR-UNQ ($\underline{\mathcal{M}3}$); QINCo ($\underline{\mathcal{M}4}$); STLQ ($\mathcal{M}5$). The underlines represent that data are cited}

    \label{tab_recall_1m_datasets}
    \begin{threeparttable}
    \begin{tabularx}{\textwidth}{m{9pt} XXXXXX | XXXXXX | XXXXXX }
        \toprule
        & \multicolumn{3}{c}{\bf SIFT1M-8B}   & \multicolumn{3}{c}{\bf SIFT1M-16B}
        & \multicolumn{3}{c}{\bf GIST1M-8B}   & \multicolumn{3}{c}{\bf GIST1M-16B}
        & \multicolumn{3}{c}{\bf DEEP1M-8B}   & \multicolumn{3}{c}{\bf DEEP1M-16B}\\
        \cmidrule(lr){2-4} \cmidrule(lr){5-7}
        \cmidrule(lr){8-10} \cmidrule(lr){11-13}
        \cmidrule(lr){14-16} \cmidrule(lr){17-19}
        & $\mathcal{R}1$ &$\mathcal{R}10$ &$\mathcal{R}100$ &  $\mathcal{R}1$ &$\mathcal{R}10$ &$\mathcal{R}100$
        & $\mathcal{R}1$ &$\mathcal{R}10$ &$\mathcal{R}100$ &  $\mathcal{R}1$ &$\mathcal{R}10$ &$\mathcal{R}100$
        & $\mathcal{R}1$ &$\mathcal{R}10$ &$\mathcal{R}100$ &  $\mathcal{R}1$ &$\mathcal{R}10$ &$\mathcal{R}100$\\
        \midrule
        $\mathcal{M}1$
        & 29.85 & 72.75 & 97.09 & 47.97 & 92.29 & 99.96
        & 14.10 & 39.30 & 78.20 & 23.60 & 62.50 & {\bf93.50}
        & 23.10 & 66.95 & 95.54 & 43.13 & 90.21 & 99.87\\
        $\mathcal{M}2$ 
        &35.31 & 79.44 & 98.66 & 50.60 & 91.95 & 99.63
        &15.30 & 42.50 & 81.30 & 23.60 & 60.10 & 90.50
        &28.11 & 73.62 & 97.68 & 42.90 & 90.02 & 99.48\\
        $\underline{\mathcal{M}3}$ &/ & / & / & / &/ & /
        &/ & / & / & / &/ & /   
        &\textbf{32.90} & \textbf{79.90} & 93.20 &49.10 &92.90  & 97.00\\
        $\underline{\mathcal{M}4}$\tnote{*} & 35.80 &  80.40 &  98.60  & /& /& / 
        &/ & / & / & / &/ & /  
        & 29.40 &  77.60 &  \textbf{98.50}  & \textbf{53.00} &  \textbf{96.20} &  \textbf{100.0}\\
        $\mathcal{M}5$
        & \textbf{39.44} & \textbf{84.03} & \textbf{99.02} & \textbf{59.34}  & \textbf{96.88} & \textbf{99.98}
        & \textbf{18.20} & \textbf{48.80}   & \textbf{84.40}  & \textbf{28.70}   & \textbf{65.60}  & 93.10
        & 31.28 & 76.87 & 98.02 & 51.67  & 95.04  & 99.94\\
        \bottomrule
    \end{tabularx}
    \begin{tablenotes}
      \item[] STLQ topology parameters of train set and base set: SIFT1M-8B: (10,6,10), (14,12,30). 16B: (10,6,15), (12,11,40); GIST1M-8B: (10,6,15), (12,11,35). 16B: (14,10,10), (16,15,30); DEEP1M-8B: (10,6,15), (12,11,40). 16B: (10,6,15), (12,11,40).
      \item[*] QINCo uses L4 version for SIFT1M, and uses L12 version for DEEP1M. For DEEP1M, QINCo can achieve higher recall with 10M train set instead of standardized 500k here. In addition, QINCo2 reports R@1 of 45.1\% and 67.1\% on Deep1M at 8B and 16B, respectively, using 358M train set \cite{vallaeys2025qinco2}.
    \end{tablenotes}
\end{threeparttable}
\end{table*}

\begin{table}[!htb]
    \centering
    \caption{The maximum top-1 R@k on BigANN. The corresponding recall-QPS curves are reported in Fig. \ref{fig_recall_qps}}
    \label{tab_recall_bigann}
    \begin{threeparttable}
    \begin{tabular}{lcccccc}
        \toprule
        & \multicolumn{3}{c}{\bf BigANN-8B}   & \multicolumn{3}{c}{\bf BigANN-16B}\\
        \cmidrule(lr){2-4} \cmidrule(lr){5-7}
        & R@1 &R@10 &R@100 & R@1 & R@10 &R@100\\
        \midrule
        OPQ & 15.18 & 49.27 & 83.74 &
        34.55 & 80.78  & 96.68\\
        LSQ++ &18.58 & 56.17 & 88.14
        & 33.00 &76.14 & 93.27 \\
        DR-UNQ & /& /& /
        &\underline{45.80} & \underline{85.90}  & \underline{90.30}\\
        QINCo & \textbf{34.82} &  \textbf{79.34} &  92.30
        & \textbf{58.58} &  \textbf{95.20} &  97.08\\
        $\text{STLQ}^{1}$ & 26.21 & 68.97 & \textbf{94.12}
        & 47.26  & 91.60 & \textbf{99.37}\\
        $\text{STLQ}^{2}$ & 25.29 & 68.42 & 93.99
        & 48.29  & 90.95 & 98.94\\
        \bottomrule
    \end{tabular}
    \begin{tablenotes}
      \item[a] QINCo uses L4 version for 8B, and L2 version for 16B.
      \item[b] $\text{STLQ}^{1}$ uses IVF65k-30M-LL for 8B, uses IVF100k-50M-SS for 16B; $\text{STLQ}^{2}$ uses IVF100k-30M-LL for 8B, uses IVF100k-30M-SL for 16B.
    \end{tablenotes}
\end{threeparttable}
\end{table}

\begin{figure}[!htb]
  \centering
  \begin{minipage}[b]{1.0\linewidth}
  \subfloat[]{\includegraphics[width=0.48\linewidth]{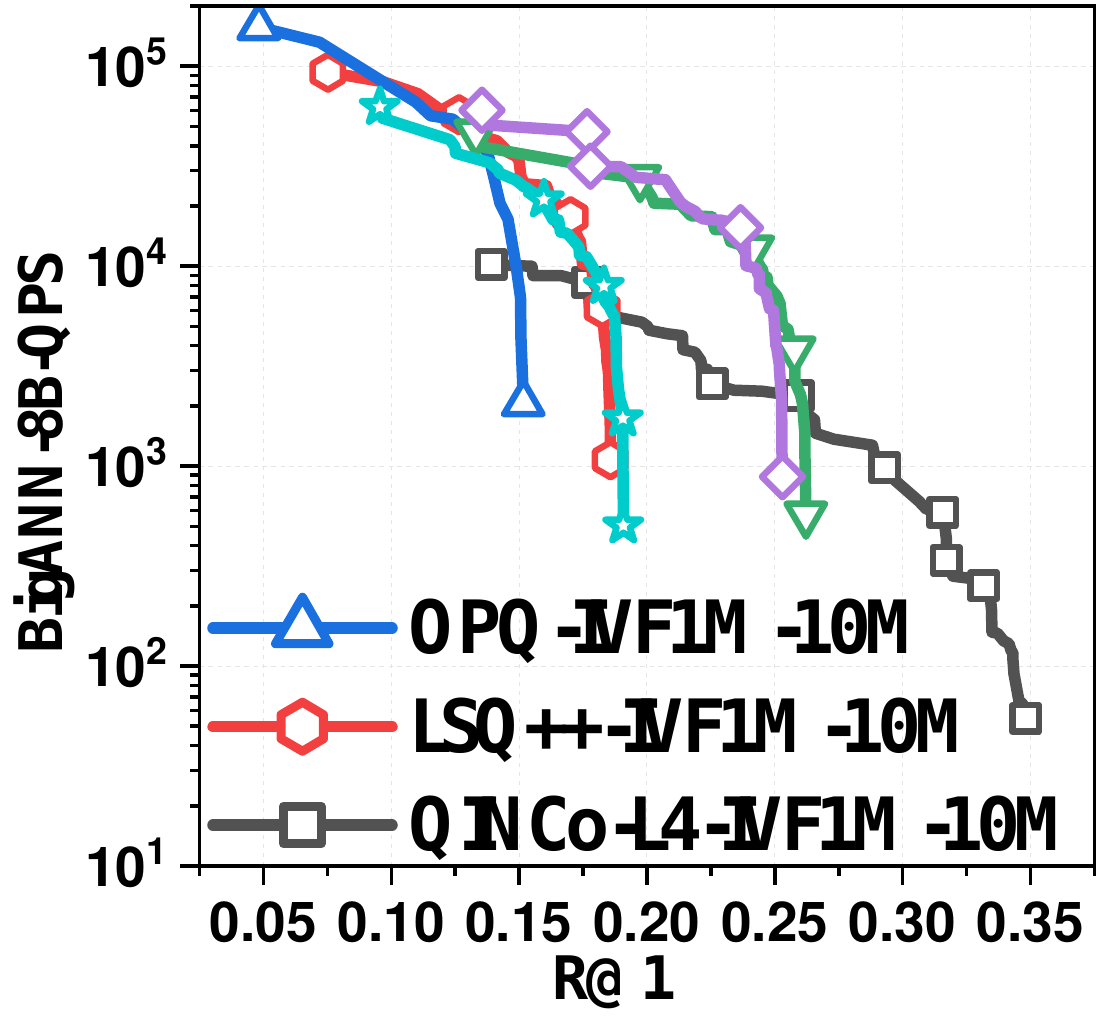}
  \includegraphics[width=0.48\linewidth]{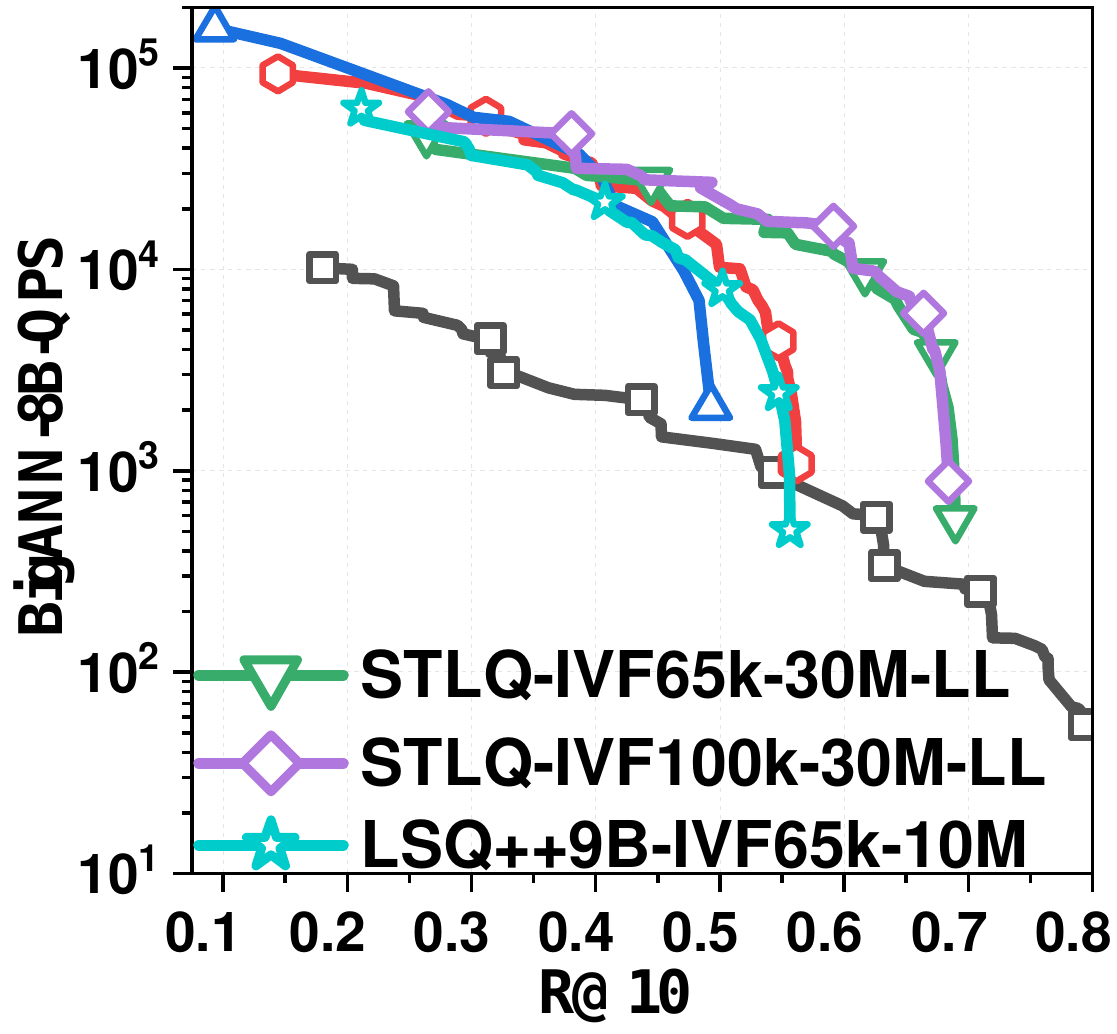}}
  \\
  \subfloat[]{\includegraphics[width=0.48\linewidth]{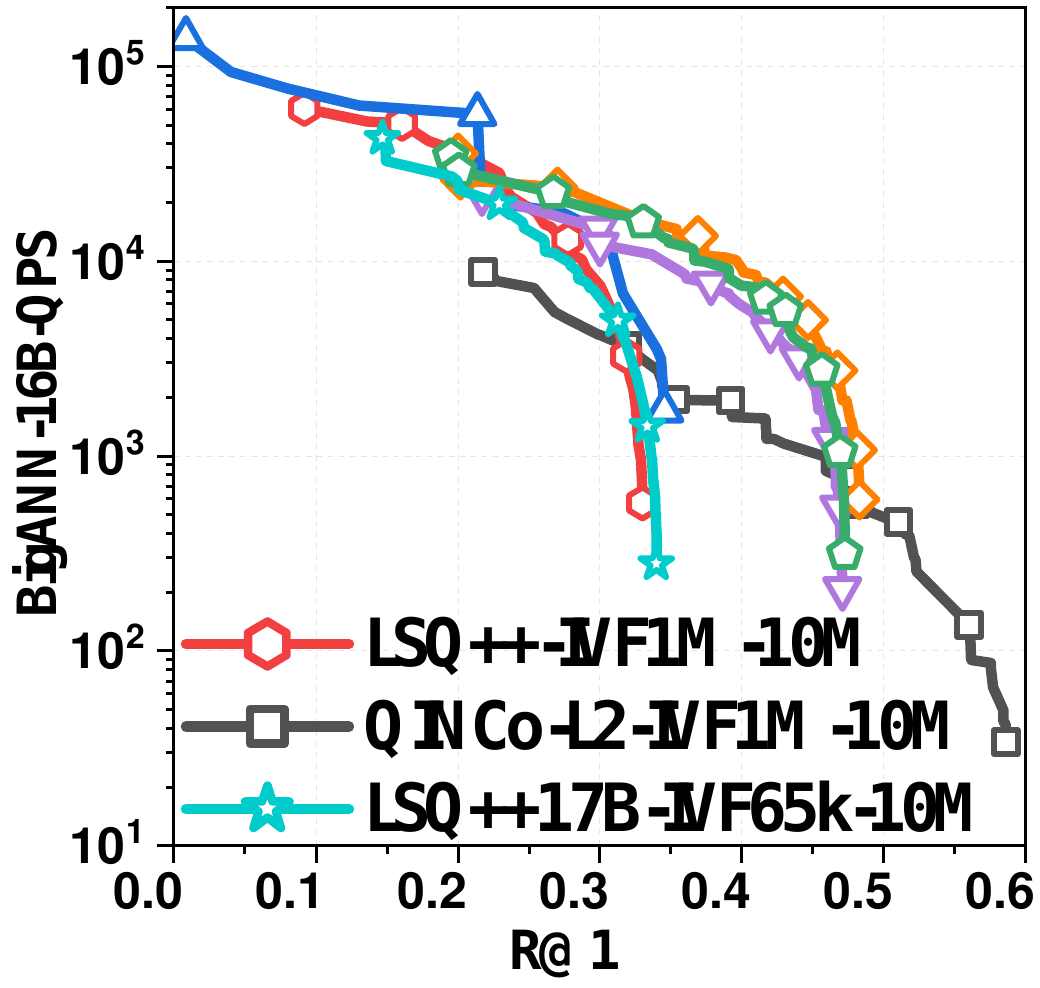}
  \includegraphics[width=0.48\linewidth]{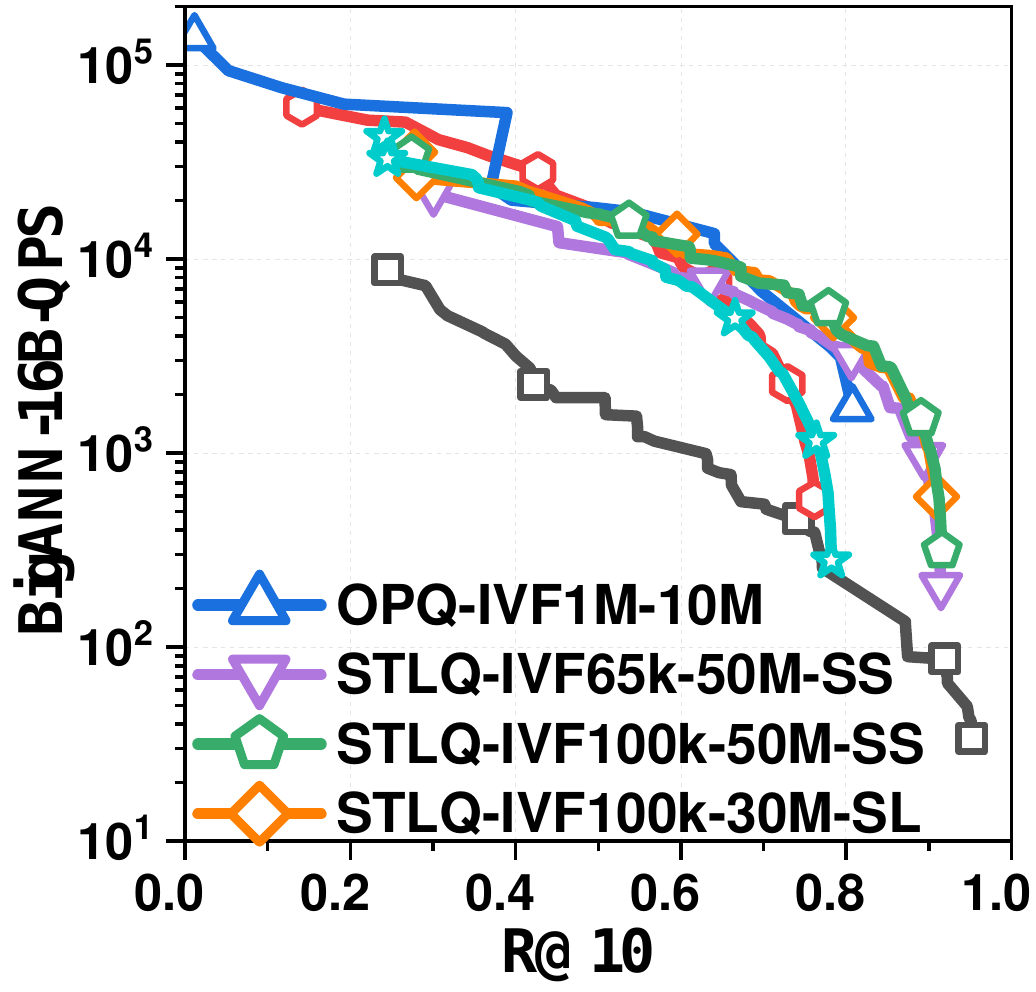}}
  \end{minipage}
  \caption{The Recall-QPS curves with 32 CPU threads on (a) BigANN-8B and (b) BigANN-16B of $\beta_p$, arranged by R@1 Pareto-optimal. Additional LSQ++-9B and 17B of matched $\beta_t$ results are explicitly noted in legends}
  \label{fig_recall_qps}
\end{figure}

\subsubsection{Recall-QPS Tradeoff}
\label{lab_recall_qps_tradeoff}
Fig. \ref{fig_recall_qps} compares the recall-QPS tradeoff on BigANN under nominal 8B and 16B of $\beta_p$. LSQ++ with 9B and 17B of $\beta_p$ is additionally evaluated to approximately match the $\beta_t$ of STLQ. All reproduced methods are evaluated using 32 threads. The QPS includes IVF probe selection, LUT construction, ADC scanning, and top-$k$ selection, while excluding index loading. Only QINCo uses compacted vectors for reranking in its standard retrieval pipeline, and the cost is included in its QPS.

\paragraph{Comparison with Shallow MCQs}
OPQ provides the highest QPS in the low recall region, partly because the Faiss implementation applies Hamming-threshold pruning to reduce unnecessary ADC evaluations. STLQ performs additional coefficient rescaling and parent-score accumulation during linked-ADC, therefore decreasing the QPS. However, OPQ and LSQ++ saturate at lower R@1. In summary, STLQ trades some QPS for higher retrieval accuracy in the medium and high recall regions.

\paragraph{Comparison with Deep MCQs}
Although QINCo reaches the highest recall, its QPS decreases a lot with longer shortlists to return more top-$k$ results, requiring more neural decoding operations. In contrast, STLQ maintains a non-neural ADC scanning path to provide higher QPS in the medium and high recall regions. DR-UNQ is compared only as a reference, because it is not open-source, and uses an NVIDIA Tesla V100 GPU for scanning in its paper. The only point in its paper achieves 45.80 R@1 with 53 QPS at 16B. In summary, OPQ favors maximum QPS at relatively low recall, while QINCo favors maximum recall at lower QPS. STLQ occupies the intermediate region, which exceeds the recall of the shallow baselines, while retaining substantially higher QPS than deep MCQs before recall saturation.

\subsubsection{Retrieval Efficiency and Native LOUDS Scanning}
\begin{figure}[!htb]
  \centering
  \begin{minipage}[b]{1.0\linewidth}
  \subfloat[]{\includegraphics[width=0.47\linewidth]{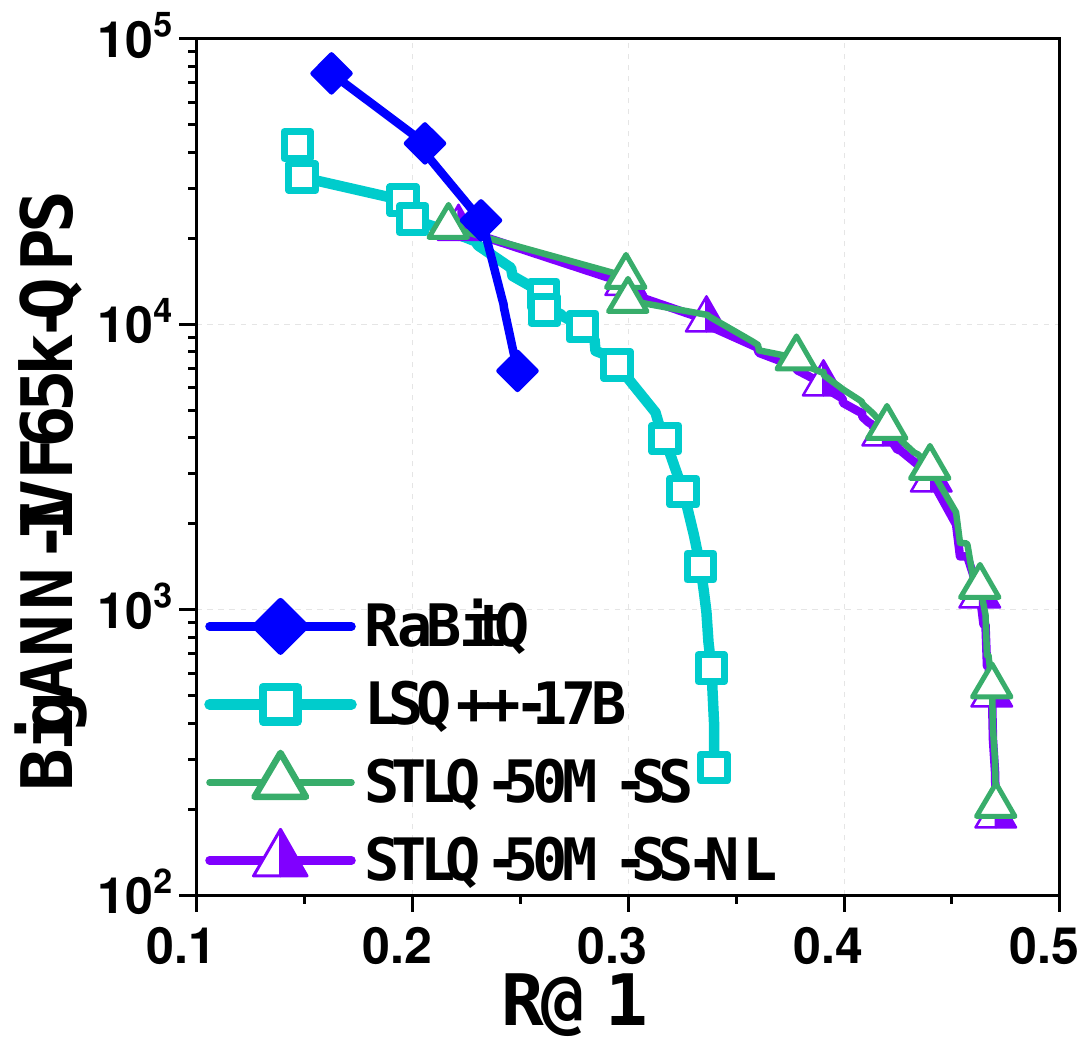}}
  \subfloat[]{\includegraphics[width=0.49\linewidth]{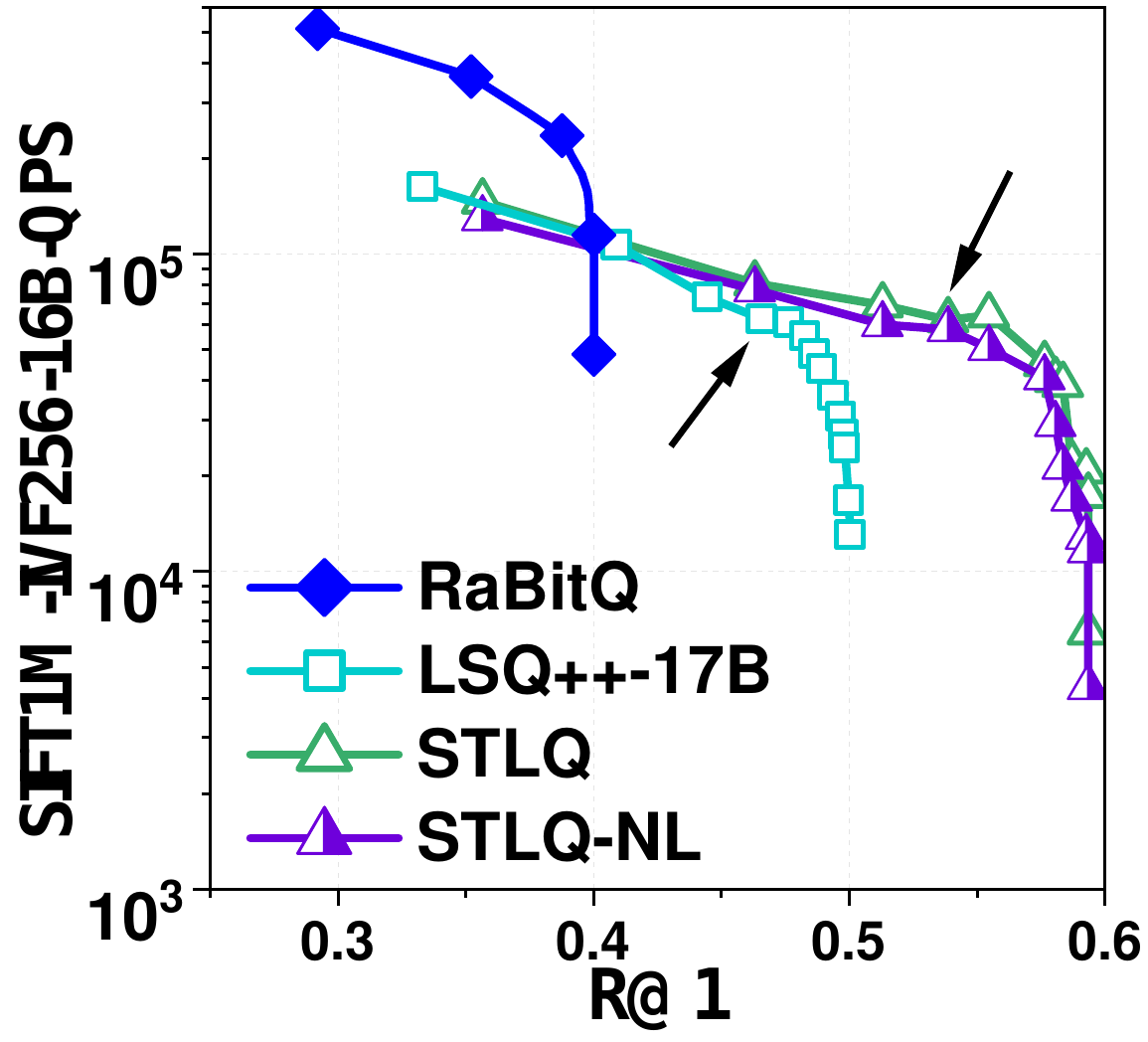}}
  \end{minipage}
  \caption{Recall-QPS (sampled markers, 32 threads) comparison with STLQ native LOUDS (NL) mode on (a) BigANN-IVF65k. (b) SIFT1M-IVF256.}
  \label{fig_qps_native_louds}
\end{figure}

Fig. \ref{fig_qps_native_louds} compares the retrieval efficiency of STLQ with LSQ++ and RaBitQ, and further evaluates native LOUDS scanning. Table \ref{tab_sift1m_cost_compare} reports the corresponding costs. Although linked-ADC introduces coefficient multiplications and parent dependencies, STLQ remains comparable to LSQ++ in QPS. This is mainly because STLQ uses fewer codebooks ($M=5/10$ versus $9/17$ for LSQ++), reducing random LUT accesses and offsetting the additional computation. STLQ maintains separate codebooks for root and linked nodes, but the two node types are scanned sequentially, so each vector accesses only one codebook set. The two sets mainly increase LUT construction cost, consistent with the slightly higher Build-LUT time in Table \ref{tab_sift1m_eval_time_breakdown}, where the Route means IVF probe selection. STLQ additionally rotates the query before scanning, leading to a slightly higher routing time. RaBitQ is also included as a complementary data-oblivious reference. It provides higher QPS but with a larger storage and lower recall than STLQ without reranking.

Compared with materializing parent array $\mathcal{P}$, native LOUDS mode sequentially decodes the bitvector in batch by efficient bit operations, saving memory at a small QPS cost. Since the topology remains unchanged, the two modes achieve identical recall. On SIFT1M, the materialized $\mathcal{P}$ can be stored in high-speed CPU cache, so that the QPS decrease of native LOUDS is slightly noticeable. While on BigANN, the smaller memory usage of native LOUDS (around 250 MB) offsets most of the decoding overhead, making the QPS decrease less obvious.

\begin{table}[!htb]
    \centering
    \caption{Quantization cost of evaluated methods in Fig \ref{fig_qps_native_louds}. RaBitQ ($\underline{\mathcal{M}1}$); LSQ++-17B ($\mathcal{M}2$); STLQ ($\mathcal{M}3$)}
    \label{tab_sift1m_cost_compare}
    \begin{threeparttable}
    \begin{tabular}{@{} ll | ccccc}
        \toprule
        && $\beta_p /bit$ & $\beta_t/bit$ & Disk /MB & Train /s & Base /s\\
        \midrule
        \multirow{3}{*}{\makecell{BigANN\\-IVF65k}}&$\underline{\mathcal{M}1}$\tnote{*}  & 192  & 208   & 26771.7  & 339 & 1140\\
        &$\underline{\mathcal{M}2}$ & 136 & 160  & 21065.0  & 1622 & 16425\\
        &$\underline{\mathcal{M}3}$ & 129.70 & 160.30  & 21153.8  & 62207 & 49935\\
        \midrule
        \multirow{3}{*}{\makecell{SIFT1M\\-IVF256}}&$\underline{\mathcal{M}1}$\tnote{*}  & 192  & 200   & 26.93  & 1.12 & 1.41\\
        &$\underline{\mathcal{M}2}$ & 136 & 152  & 23.23  & 42.27 & 16.72\\
        &$\underline{\mathcal{M}3}$    & 129.67 & 153.83  & 23.90  & 180.12 & 93.05\\
        \bottomrule
    \end{tabular}
      \begin{tablenotes}
      \item[*] For RaBitQ, $\beta_p=d+64$, accounting for its $d$-bit binary code and two float auxiliary values, while $\beta_t$ further includes the IVF overhead.
    \end{tablenotes}
    \end{threeparttable}
\end{table}

\begin{table}[!htb]
    \centering
    \caption{Average time breakdown ($/\mu$s) of single query with single CPU thread retrieval on SIFT1M-IVF256, using the points marked by arrows in Fig \ref{fig_qps_native_louds} (b), whose number of probes are both 4}
    \label{tab_sift1m_eval_time_breakdown}
    \begin{tabular}{lcccccc}
        \toprule
        & QPS & Route & \makecell{Build\\LUT} & Scan & Top-$k$ & \makecell{R@1\\/\%}\\
        \midrule
        LSQ++-17B &3809 & 1.431 & 6.317  & 208.582  & 46.322 &46.58\\
        STLQ    &3999 & 1.608 & 7.114  & 191.692  & 49.628 &53.85 \\
        \bottomrule
    \end{tabular}
\end{table}

\subsubsection{Parameter Sensitivity}
\label{lab_parameter_sensitivity}
As MCQ methods are data-dependent, quantization quality varies with hyperparameter settings and random seeds. Table \ref{tab_recall_sensitivity} reports recall statistics over 20 end-to-end runs with random seeds and fixed topology hyperparameters. The total bit budget $\beta_t$ of STLQ is approximately matched to LSQ++. In most cases, STLQ uses lower $\beta_t$ than LSQ++ to achieve a better recall distribution, except for SIFT1M, where STLQ uses 1.78 more bits to improve R@1 with 8.12, indicating a better rate-recall tradeoff.

\begin{figure}[!htb]
  \centering
  \begin{minipage}[b]{1.0\linewidth}
  \subfloat[]{\includegraphics[width=0.49\linewidth]{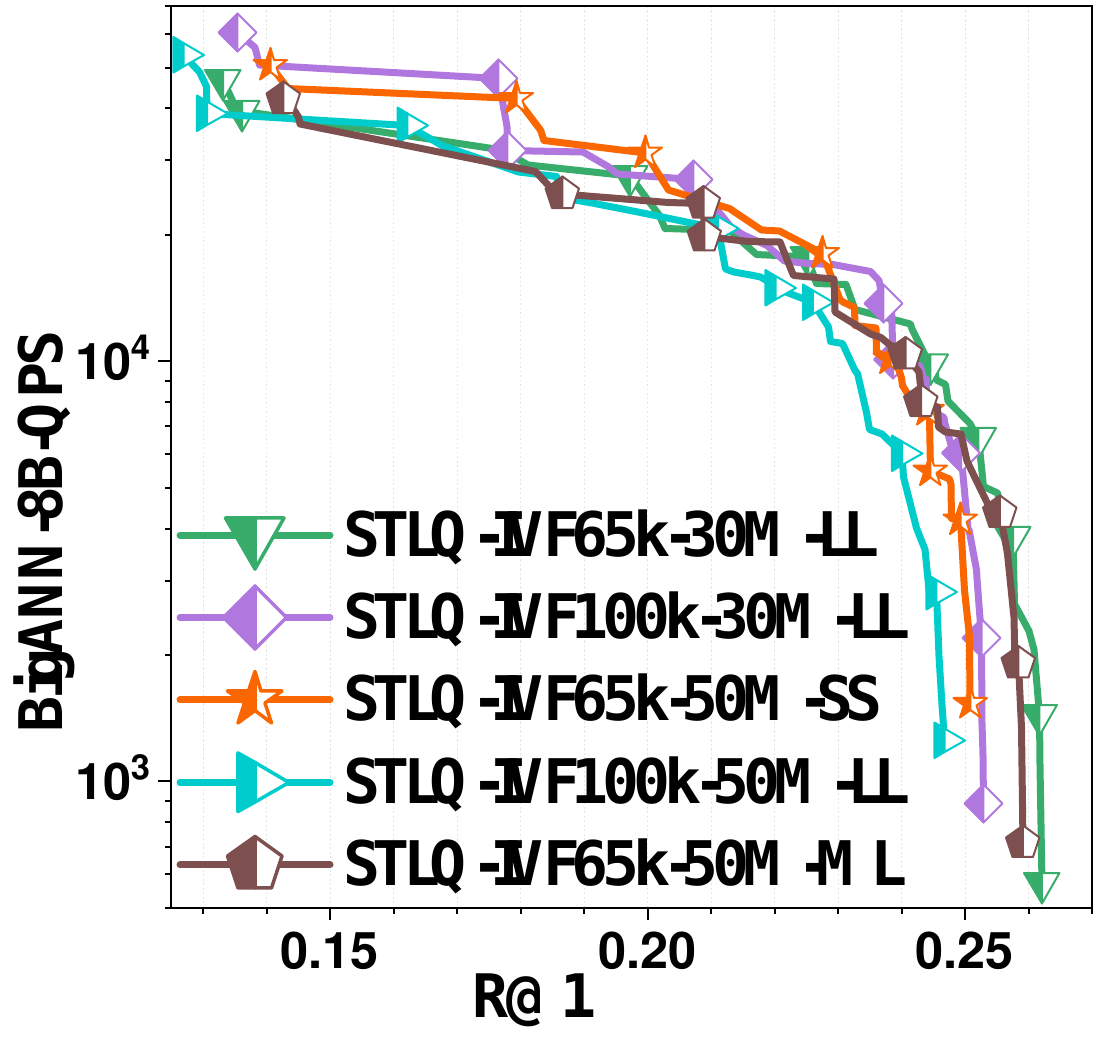}}
  \subfloat[]{\includegraphics[width=0.49\linewidth]{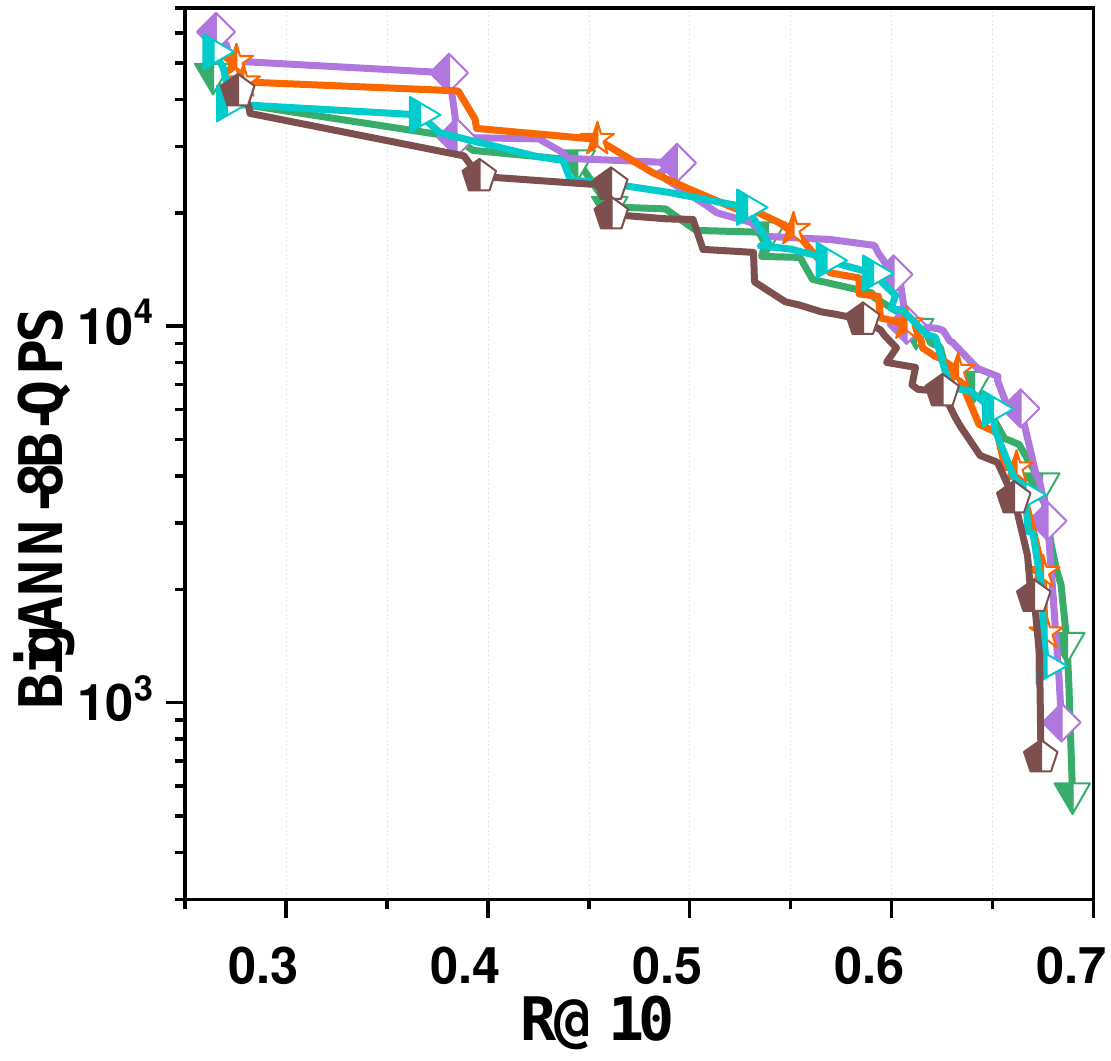}}
  \end{minipage}
  \caption{The Recall-QPS curves (sampled markers) of STLQ-8B with 32 CPU threads on BigANN. (a) R@1. (b) R@10.}
  \label{fig_stlq_bigann_b8_params}
\end{figure}

Fig. \ref{fig_stlq_bigann_b8_params} presents more recall-QPS curves of STLQ-8B than Fig. \ref{fig_recall_qps}, showing results under more settings. Considering the QPS and recall simultaneously, STLQ-IVF65k-30M-LL is preferred for 8B, and STLQ-IVF100k-30M-SL is preferred for 16B. QINCo \cite{huijben2024residual} reports that increasing train set size provides little benefit or can be harmful for shallow MCQs. However, STLQ needs to capture the topology within a cluster, so increasing the train set from 10M to 30M or 50M generally improves retrieval accuracy.

\begin{table*}[t]
\centering
\caption{Recall sensitivity over 20 random end-to-end runs per cell. In each cell of R@1 and R@10, the first line is mean $\pm$ standard deviation, and the second line is the [$\min,\max$] range. $\beta_p,\beta_t$ denote the average payload and total bit budgets, respectively. All methods use IVF-256. STLQ ($\mathcal{M}1$) uses $M=5$ and $M=10$. LSQ++ ($\mathcal{M}2$) uses $M=9$ and $M=17$ to approximately match $\beta_t$}
\label{tab_recall_sensitivity}
\begin{tabular}{@{}m{20pt} | ccc ccc ccc@{}}
\toprule
&\multicolumn{3}{c}{\textbf{SIFT1M}} & \multicolumn{3}{c}{\textbf{GIST1M}} & \multicolumn{3}{c}{\textbf{DEEP1M}} \\
\cmidrule(lr){2-4}\cmidrule(lr){5-7}\cmidrule(lr){8-10}

&$\beta_p,\beta_t$ & \textbf{R@1} & \textbf{R@10}
&$\beta_p,\beta_t$ & \textbf{R@1} & \textbf{R@10}
&$\beta_p,\beta_t$ & \textbf{R@1} & \textbf{R@10} \\
\midrule

\makecell{$\mathcal{M}2$\\-9B}
& \makecell{72,\\88} & \makecell{$36.79\pm 0.4802$\\$[35.45,37.43]$} & \makecell{$81.20\pm0.7536$\\$[79.74,82.21]$} 
& \makecell{72,\\88} & \makecell{$15.39\pm1.138$\\$[12.50,17.10]$} & \makecell{$43.79\pm1.378$\\$[41.10,46.50]$}
& \makecell{72,\\88} & \makecell{$29.66\pm0.4976$\\$[28.31,30.31]$} & \makecell{$76.18\pm0.9356$\\$[72.80,77.07]$} \\

\cmidrule(lr){2-10}

\makecell{$\mathcal{M}1$\\-8B}
& \makecell{65.31,\\86.23} & \makecell{$38.65\pm0.4834$\\$[37.65,39.44]$} & \makecell{$83.65\pm0.3553$\\$[83.11,84.60]$}
& \makecell{64.31,\\84.61} & \makecell{$16.55\pm1.155$\\$[14.40,18.20]$} & \makecell{$46.21\pm1.210$\\$[44.00,48.80]$}
& \makecell{64.88,\\84.85} & \makecell{$30.04\pm0.4334$\\$[29.06,30.88]$} & \makecell{$76.70\pm0.3262$\\$[76.11,77.36]$} \\

\midrule

\makecell{$\mathcal{M}2$\\-17B}
& \makecell{136,\\152} & \makecell{$50.11\pm 0.7363$\\$[47.92,51.31]$} & \makecell{$92.35\pm1.318$\\$[88.38,93.79]$} 
& \makecell{136,\\152} & \makecell{$22.70\pm1.313$\\$[20.70,25.40]$} & \makecell{$59.80\pm1.701$\\$[56.60,62.20]$}
& \makecell{136,\\152} & \makecell{$43.31\pm1.374$\\$[38.61,44.69]$} & \makecell{$89.20\pm2.294$\\$[81.80,91.07]$} \\

\cmidrule(lr){2-10}

\makecell{$\mathcal{M}1$\\-16B}
& \makecell{129.74,\\153.78} & \makecell{$58.23\pm0.4564$\\$[57.44,59.08]$} & \makecell{$96.66\pm0.2160$\\$[96.30,97.09]$}
& \makecell{125.61,\\148.62} & \makecell{$24.99\pm0.8857$\\$[23.20,26.90]$} & \makecell{$65.14\pm1.140$\\$[63.10,66.80]$}
& \makecell{129.54,\\151.73} & \makecell{$50.94\pm0.4039$\\$[49.90,51.67]$} & \makecell{$94.73\pm0.1828$\\$[94.34,95.04]$} \\

\bottomrule
\end{tabular}
\end{table*}

\subsection{Ablation and Analysis}
\subsubsection{Ablations}
We present the ablations of STLQ component, coefficient compression, and topology construction. Through these studies, we show the component contribution, and highlight the importance of storage space control and topological structural constraints. Table \ref{tab_ablations} reports the ablation results on SIFT1M-16B. Removing the linked encoding phase decreases R@1, confirming that reusing the parent reconstruction improves retrieval accuracy. The virtual-relay topology further improves recall at the cost of 4.5\% higher $\beta_t$ with $\gamma_v=0.05$. Orthogonal rotation also better aligns the codebooks with the vector distribution to improve recall. Quantizing the Float32 coefficients $\alpha$ into 6-bit signed integers causes only a small recall drop while saving substantial $\beta_t$. Subsequent lossless Huffman coding further reduces the storage overhead without affecting recall. As for the parent-reference ablations, they are evaluated without virtual nodes. IVF reference only is just equal to basic encoding. Approximate nearest neighbor (ANN) node reference slightly reduces MSE and improve recall, indicating that geometric distance alone is insufficient for selecting effective parent-references. In contrast, directly using random nodes provides almost no accuracy gain, and the linked ratio $\gamma_l$ is tiny.

\begin{table}[t]
\centering
\caption{Ablations on SIFT1M-IVF256-16B}
\label{tab_ablations}
\begin{tabular}{lccccc}
\toprule
    & \makecell{MSE\\/$\cdot 10^3$} & R@1 & R@10 &\makecell{$\gamma_l$\\/\%} & \makecell{$\beta_t$\\/bit} \\
    \midrule
    \multicolumn{5}{l}{\textbf{Component Ablations}} \\
    Full STLQ
        & 7.69   & 59.34 & 96.88 & 92.21 & 153.83 \\
    w/o Linkage
        & 10.21   & 51.69 & 94.18 & 0.00 & 138.28 \\
    w/o Virtual-Relay
        & 8.34   & 55.19 & 96.04 & 82.46 & 147.20 \\
    w/o OPP rotation
        & 7.69   & 57.86 & 96.88 & 92.12 & 153.82 \\
\midrule
\multicolumn{5}{l}{\textbf{Coefficient Compression Ablations}}  \\
    Float32 $\alpha$ Linked-ADC
        & 7.59  & 59.99 & 97.11 & 92.21 & 436.92 \\
    + Lossy 6-bit integer $\alpha$
        & 7.69  & 59.34 & 96.88 & 92.21 & 164.19 \\
    + Lossless Huffman
        & 7.69  & 59.34 & 96.88 & 92.21 & 153.83 \\
\midrule
\multicolumn{5}{l}{\textbf{Parent Topological Reference Ablations}}  \\
    IVF center
        & 10.21  & 51.69 & 94.18 & 0.00 & 138.28 \\
    Random node
        & 10.21  & 51.45 & 94.30 & 0.77 & 140.36 \\
    ANN node
        & 9.04  & 53.57 & 94.95 & 46.19 & 144.10 \\
\bottomrule
\end{tabular}
\end{table}

\subsubsection{ADC Error Analysis}
\begin{figure}[!htb]
  \centering
  \begin{minipage}[b]{1.0\linewidth}
  \subfloat[]{\includegraphics[width=0.326\linewidth]{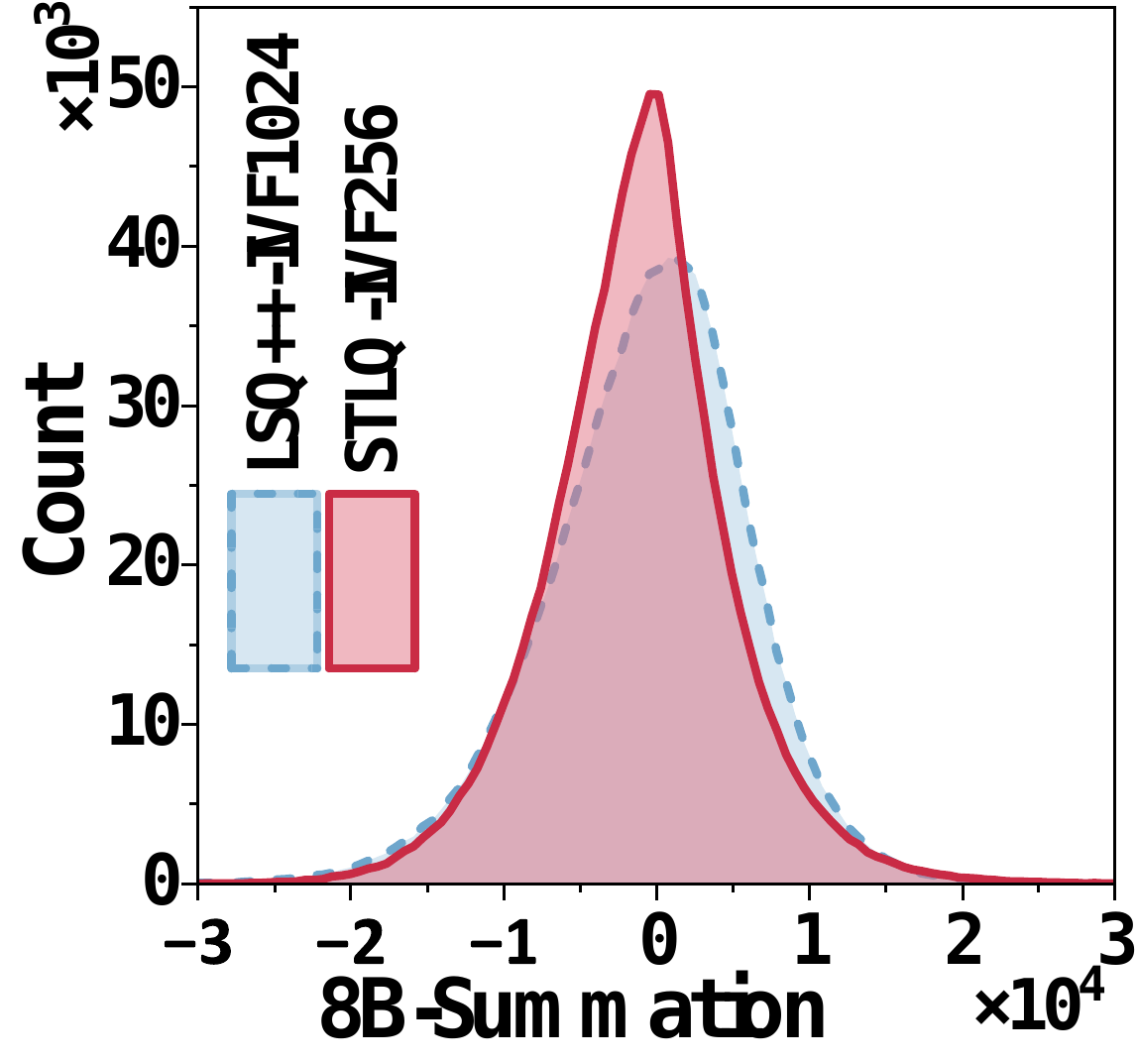}
  \includegraphics[width=0.323\linewidth]{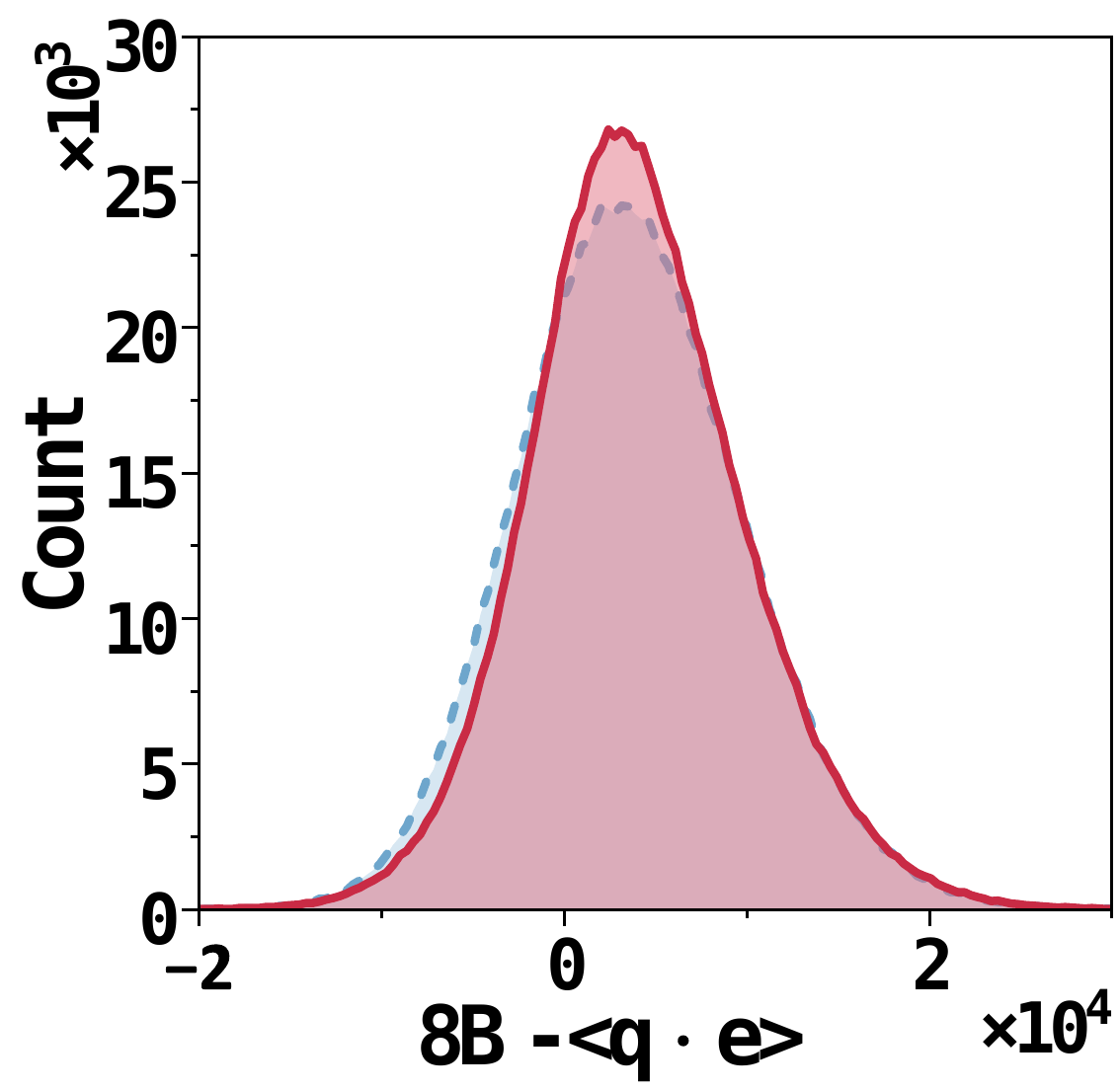}
  \includegraphics[width=0.323\linewidth]{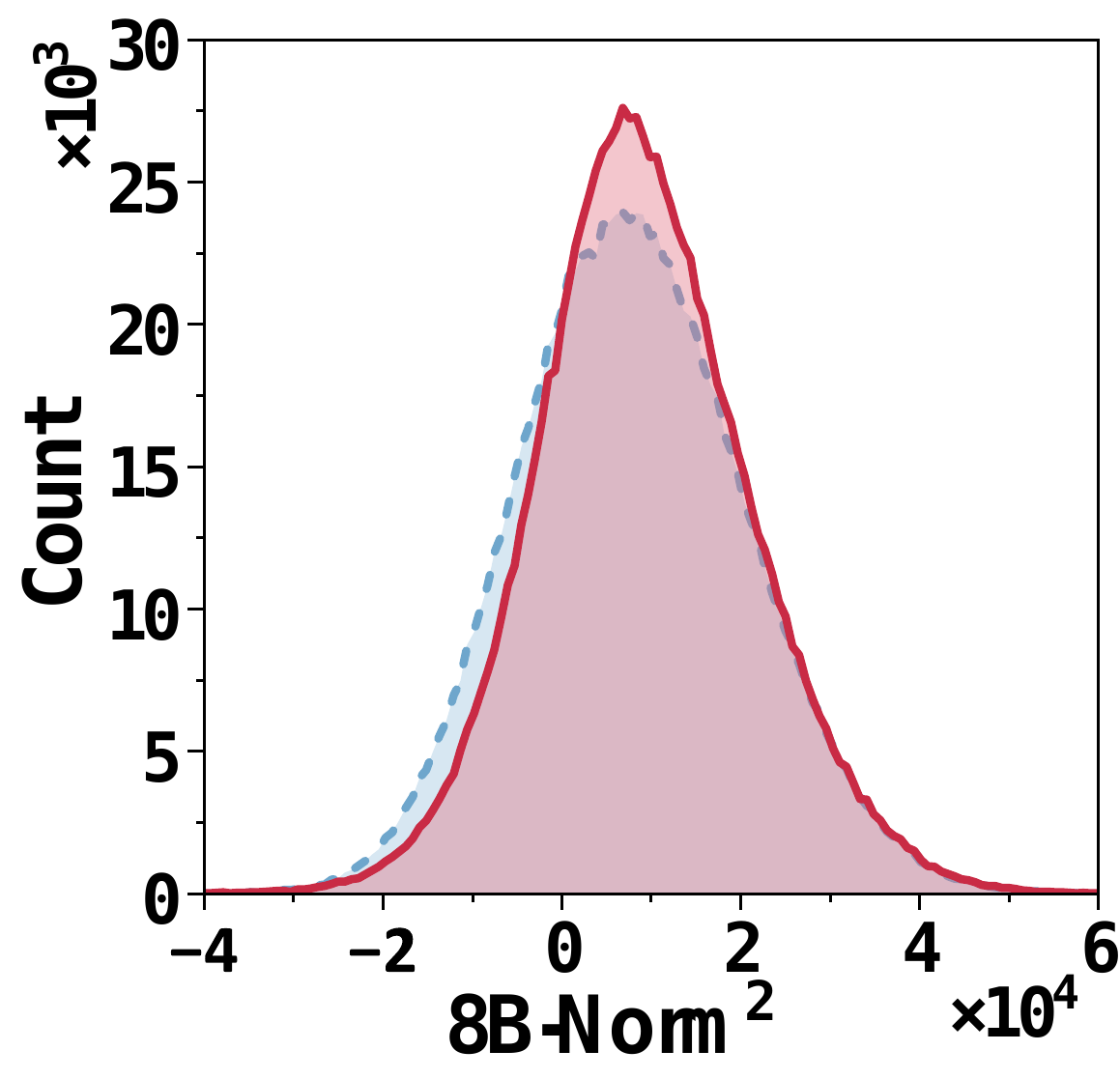}}\\
  \subfloat[]{\includegraphics[width=0.323\linewidth]{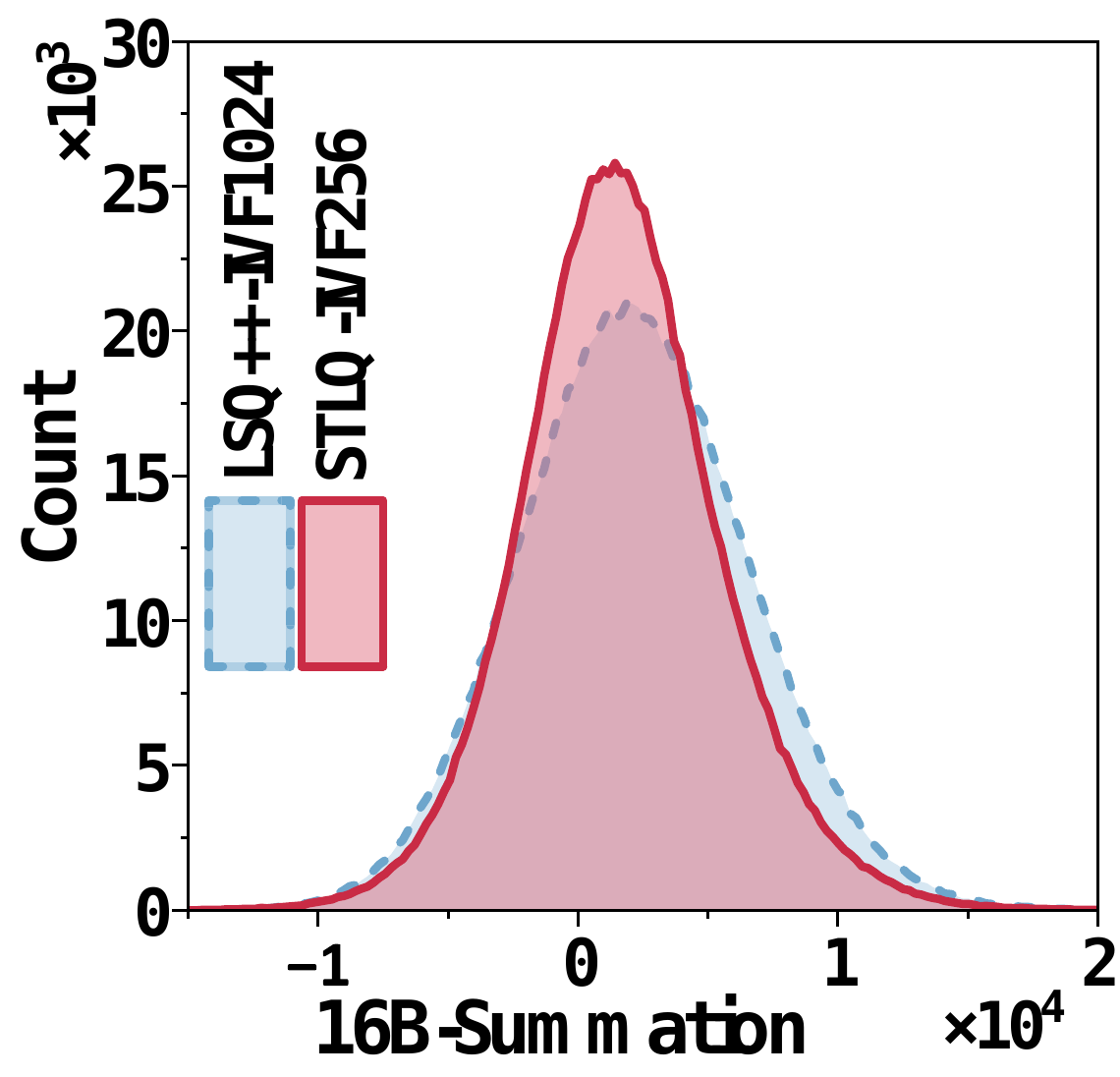}
  \includegraphics[width=0.323\linewidth]{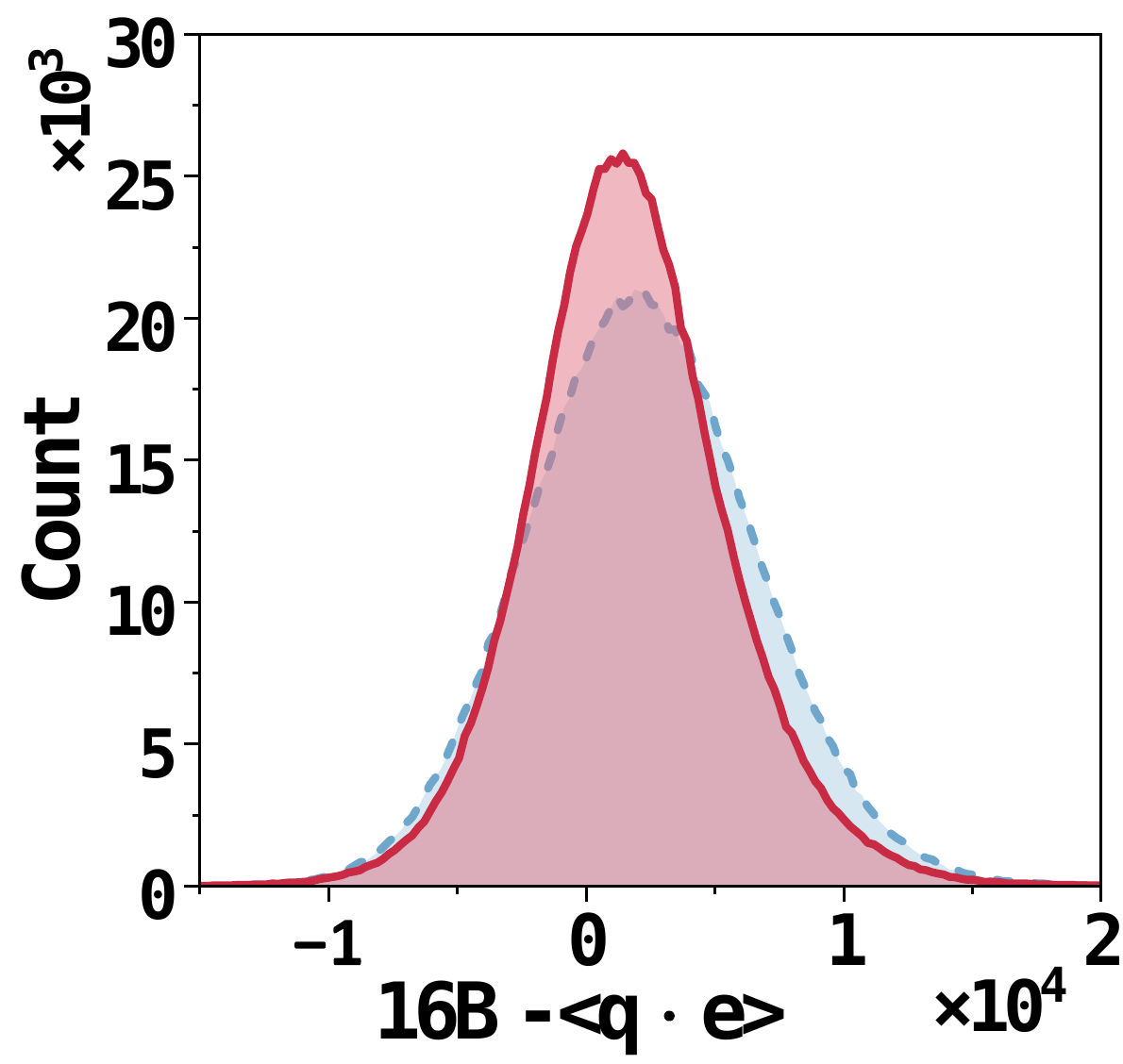}
  \includegraphics[width=0.323\linewidth]{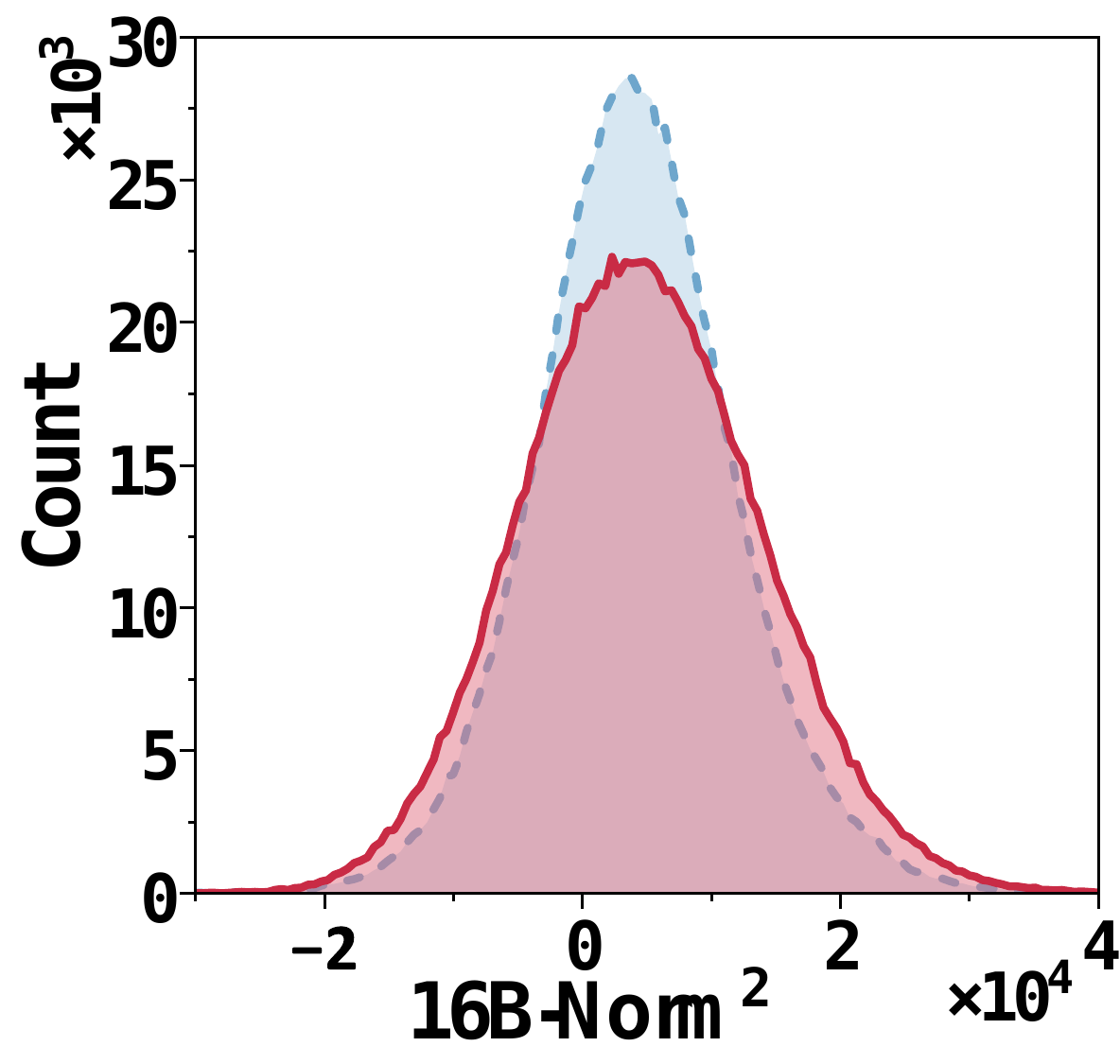}}
  \end{minipage}
    \caption{Error distribution of (a) SIFT1M-8B and (b) SIFT1M-16B in Eq. \ref{fun_diff_err} on top-100 ground truth. Figures with ``Summation'' on x-axis show the per vector sum of inner-product and norm terms}
  \label{fig_adc_compare}
\end{figure}

Fig. \ref{fig_adc_compare} shows the error distribution of STLQ and LSQ++ in Eq. \ref{fun_diff_err} on the top-100 ground truth of SIFT1M. Although these plots show the single vector errors rather than the pairwise interaction determining ranking inversions, they could explicitly reveal the empirical upper bound, where a narrower range indicates a smaller inversion possibility. For 16B quantization, LSQ++ achieves a better L2-norm distribution, while STLQ improves the dot-product term and reverses the tendency of total error, indicating linked encoding better aligns residual orientation.

\subsubsection{Topology Distribution Analysis}
\begin{figure}[!htb]
  \centering
  \begin{minipage}[b]{1.0\linewidth}
  \subfloat[]{\includegraphics[width=0.32\linewidth]{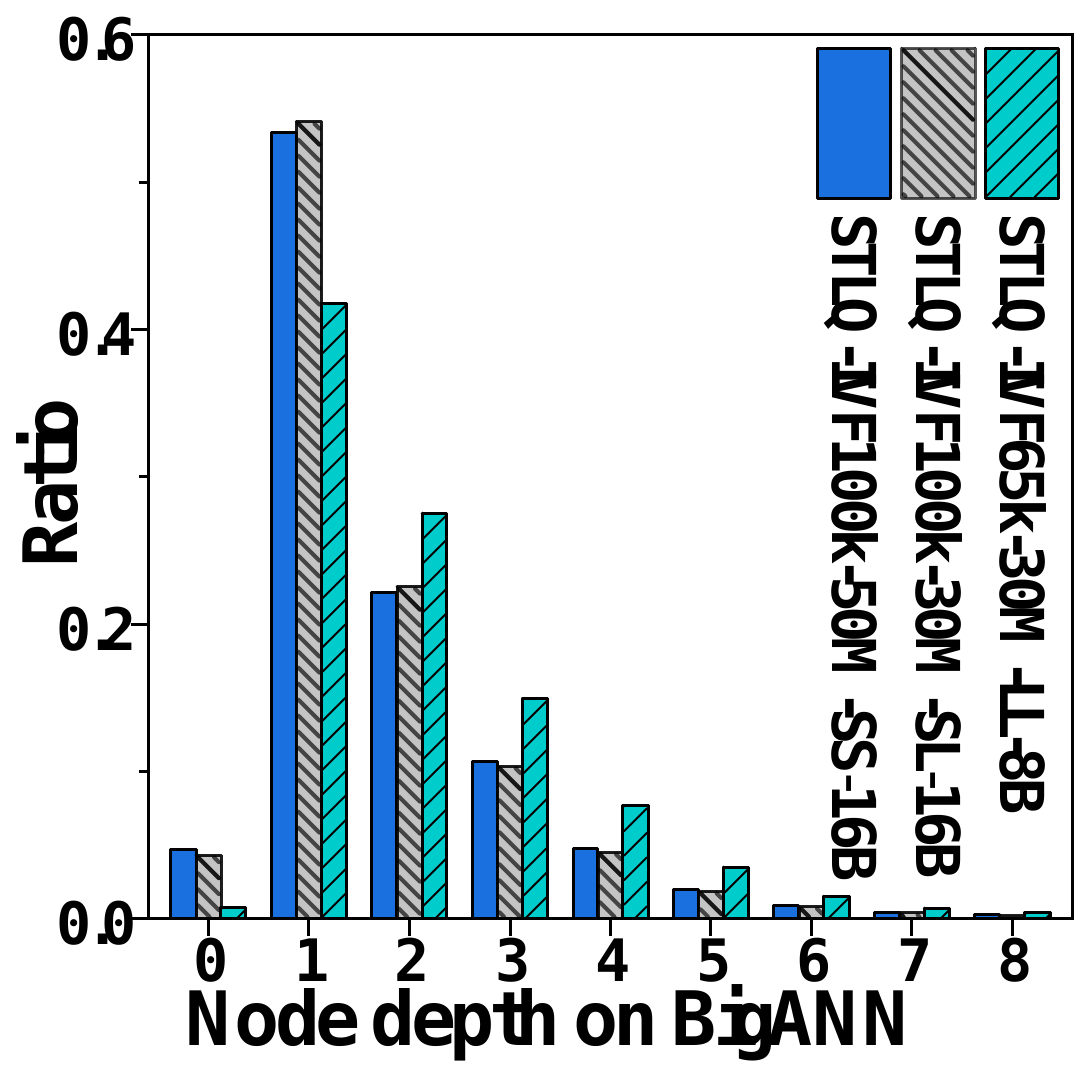}}
  \subfloat[]{\includegraphics[width=0.328\linewidth]{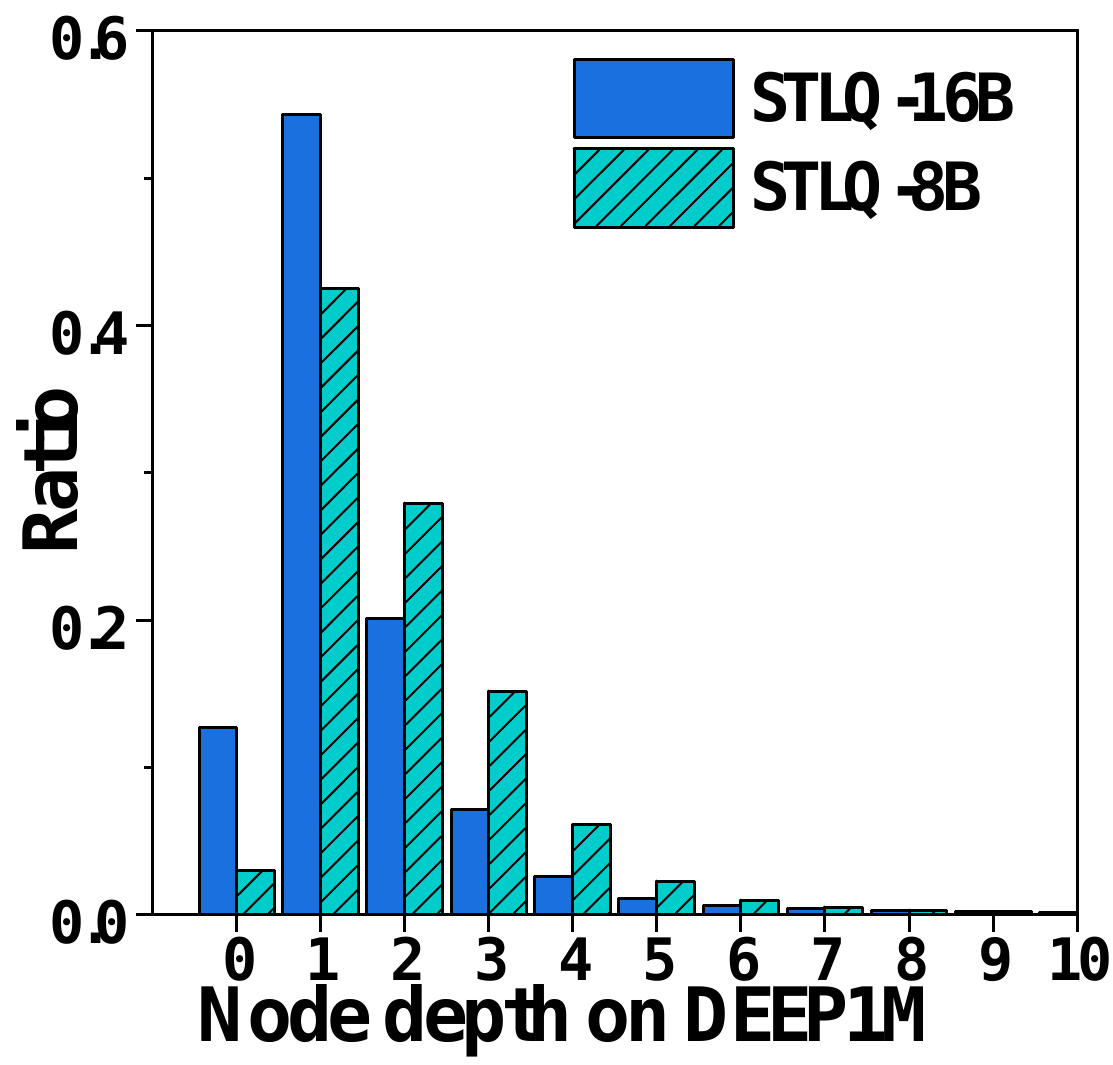}}
  \subfloat[]{\includegraphics[width=0.330\linewidth]{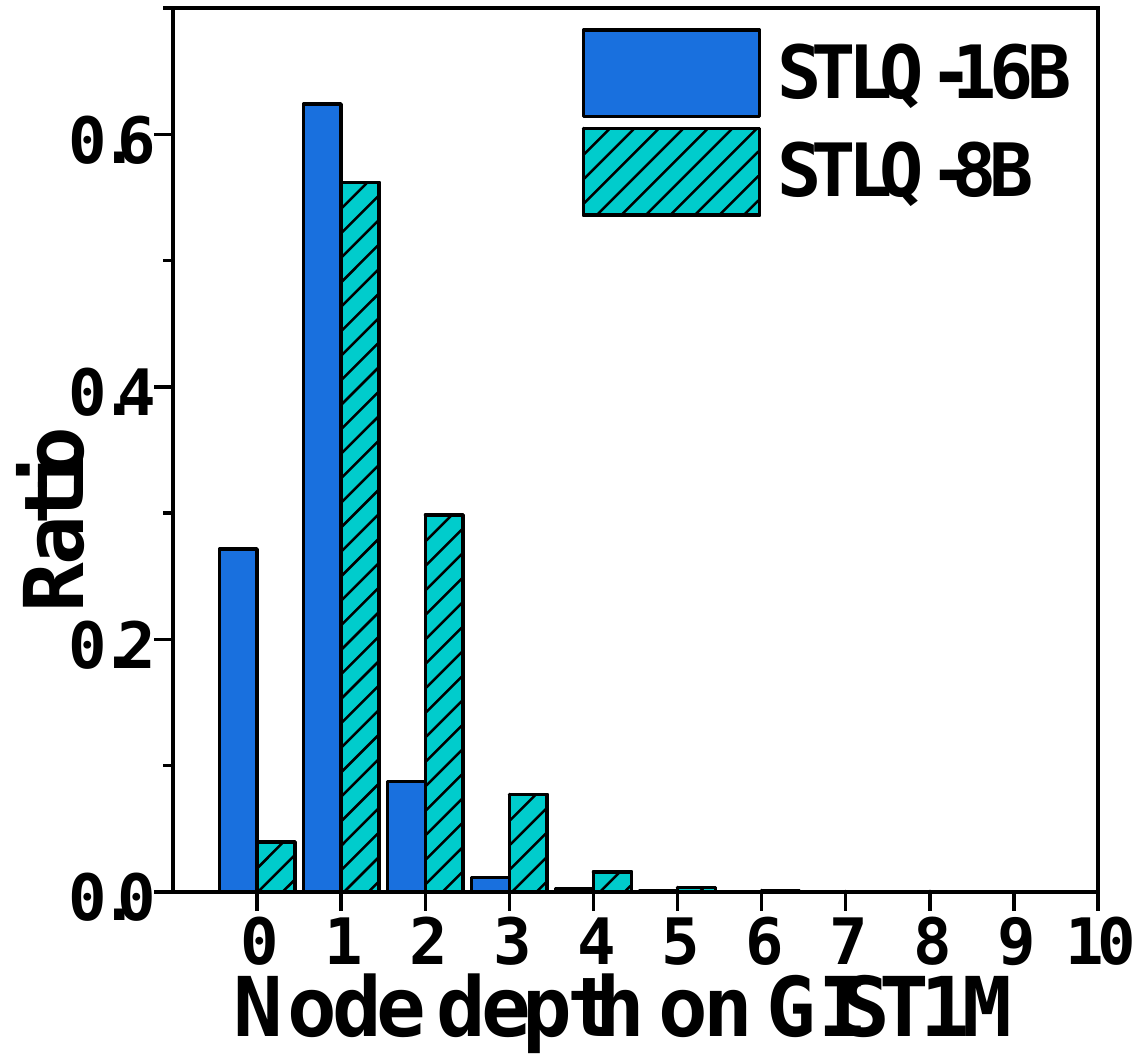}}
  \end{minipage}
    \caption{Node-depth distribution. (a) BigANN. (b) DEEP1M. (c)GIST1M.}
  \label{fig_depth_distribution}
\end{figure}

Fig. \ref{fig_depth_distribution} illustrates the node-depth distribution of STLQ. Most nodes are successfully linked, while the distribution decays rapidly with depth, and the majority of linked nodes remain within depth 5. The 16B quantization retains slightly more root nodes because its quantization quality is already good. GIST1M also exhibits a higher root ratio and shallower topology, we believe useful local references are harder to obtain with its higher dimensionality. Generally, the depth distribution reflects the utilization of local topology, denser IVF cells and larger train sets provide more useful parent candidates, and therefore tend to increase the linked ratio, which is consistent with our experiments.

\subsubsection{Reconstructed Angle Analysis}
\begin{figure}[!htb]
  \centering
  \begin{minipage}[b]{1.0\linewidth}
  \subfloat[]{\includegraphics[width=0.321\linewidth]{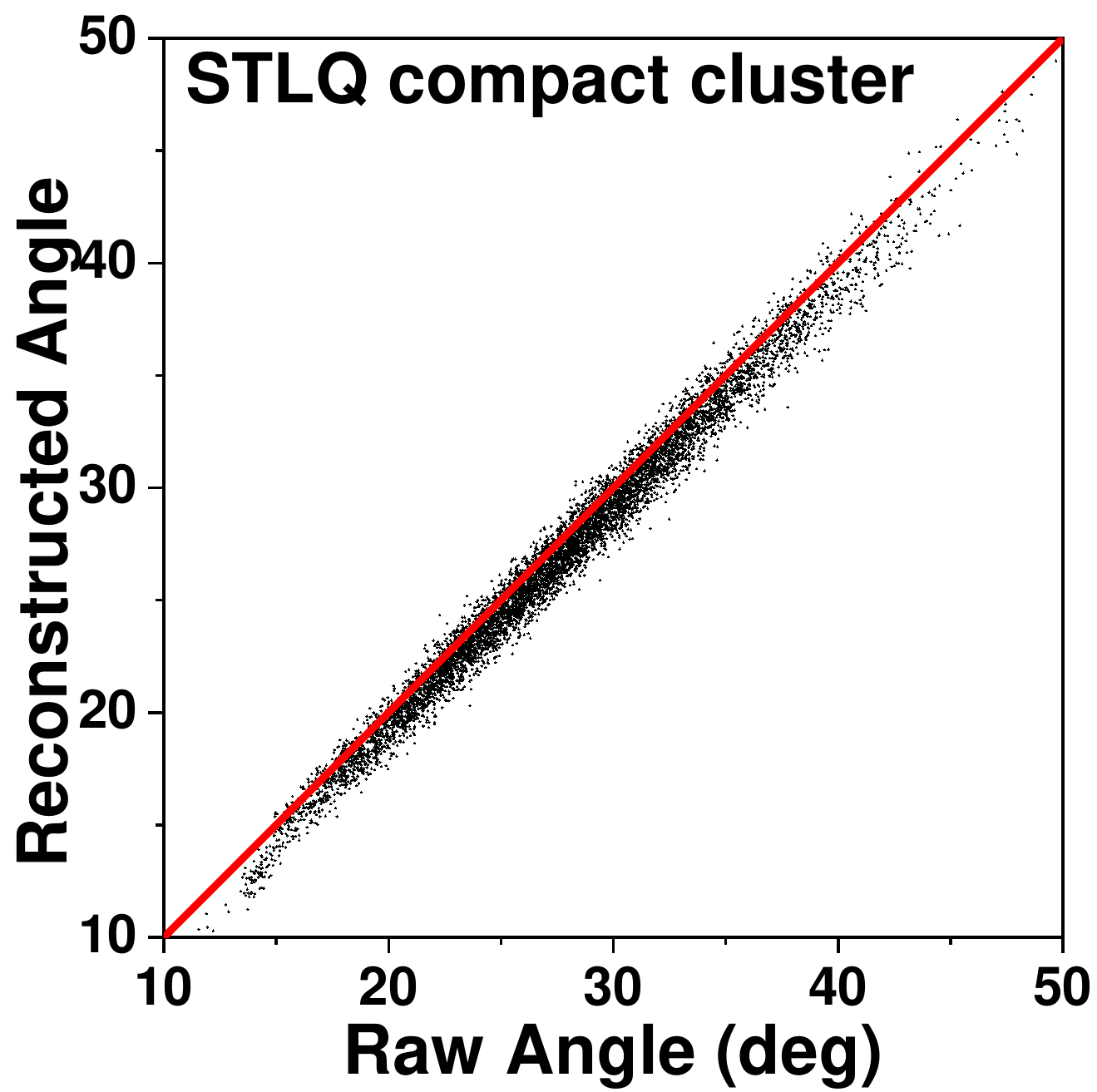}}
  \subfloat[]{\includegraphics[width=0.321\linewidth]{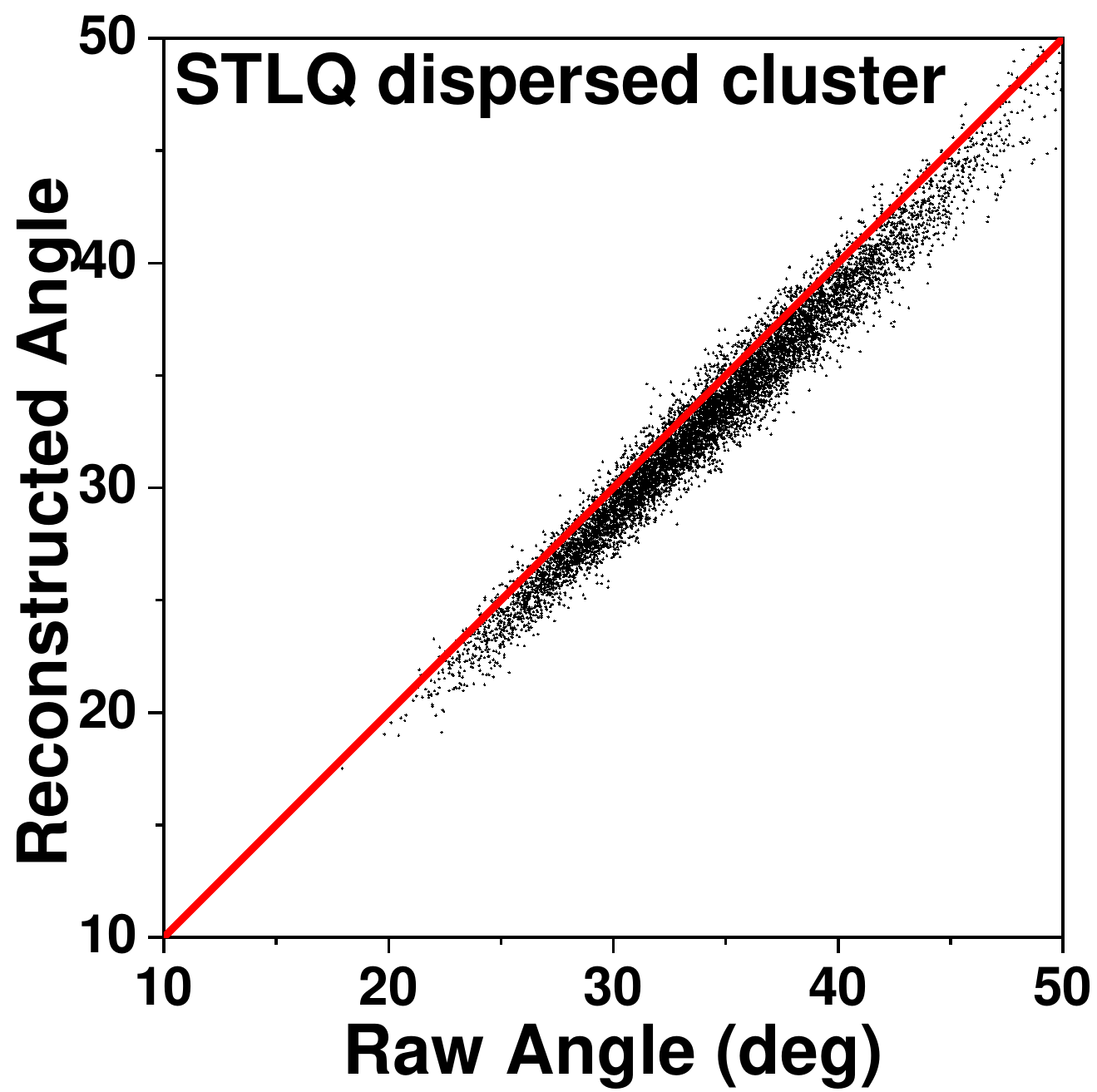}}
  \subfloat[]{\includegraphics[width=0.319\linewidth]{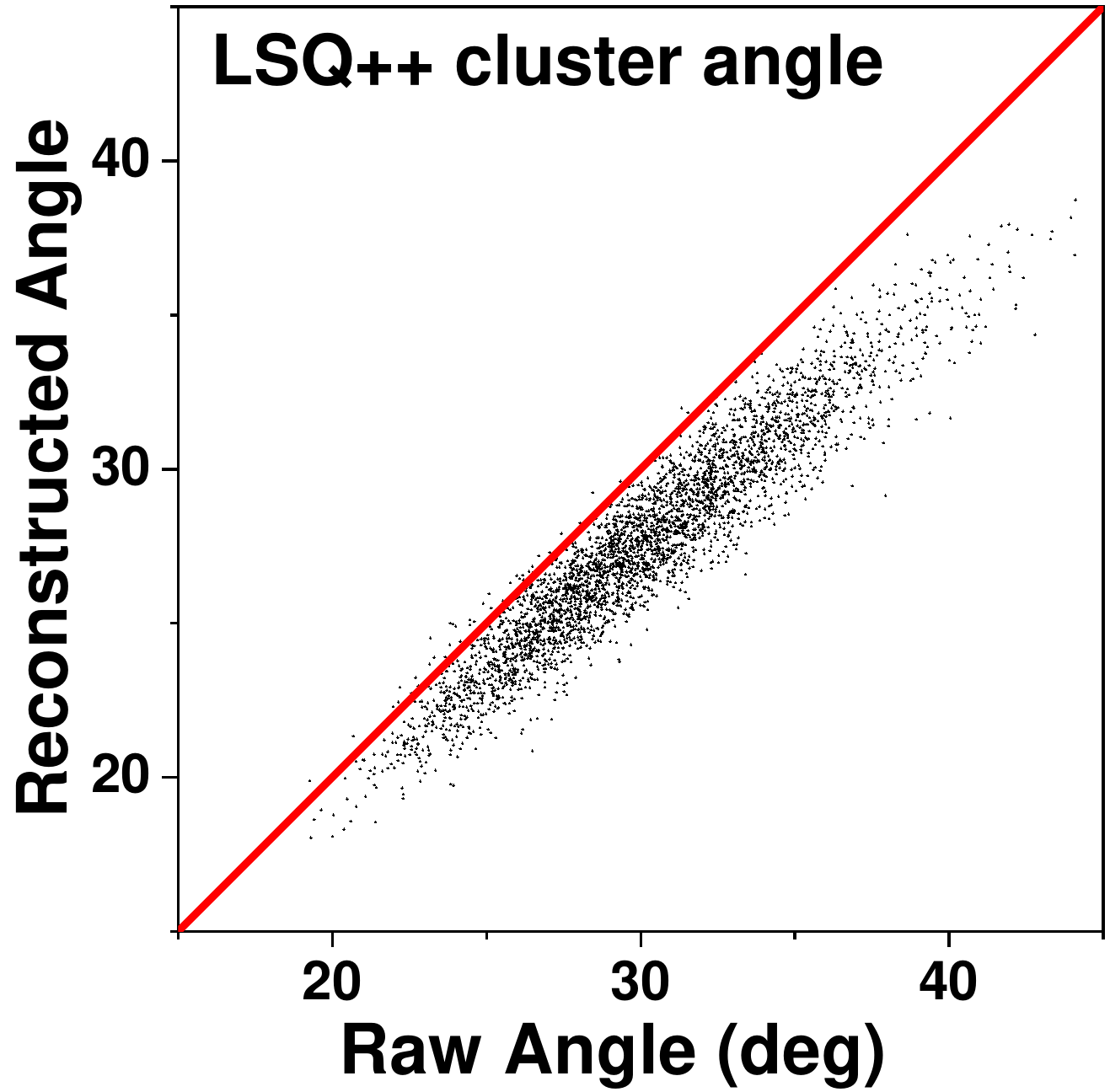}}
  \end{minipage}
  \caption{Comparison between cluster-wise reconstructed angle and raw angle. (a) STLQ compact cluster. (b)STLQ dispersed cluster. (c) LSQ++ cluster.}
  \label{fig_raw_angle_vs_reconstructed_angle}
\end{figure}

To validate our angle-deviation motivation discussed in Section \ref{sec_introduction}, Fig. \ref{fig_raw_angle_vs_reconstructed_angle} analyzes the angle deviation from the IVF centroid for STLQ and LSQ++ on SIFT1M. For both compact and dispersed clusters, the reconstructed angles of STLQ roughly follow the raw angles, with most points distributed near the diagonal. In contrast, LSQ++ exhibits an increasing angular deviation as the raw angle grows, indicating that its reconstructed vectors are more biased toward the cluster centroid, which helps explain its lower recall. In fact, the basic encoding representation of STLQ also has the same behavior as LSQ++. This result shows that the linked encoding in STLQ better preserves the original angular structure within each IVF cluster.

\section{Conclusion}
\label{lab_conclusion}
We presented STLQ, a shallow MCQ method that improves vector quantization by exploiting topological references within IVF clusters. STLQ organizes each cluster as a directed-forest and re-encodes relative residuals between linked vectors, providing finer local resolution. Linked-ADC retains linear scanning complexity, while coefficient compression and LOUDS effectively control the additional storage overhead. Extensive experiments show that STLQ improves recall over the evaluated shallow MCQs, while maintaining competitive query efficiency. The parent-reference and angle analyses further show that geometric distance alone does not guarantee an effective reference, whereas linked encoding substantially reduces reconstruction bias toward the IVF centroid. Although STLQ incurs additional offline construction cost, it provides a favorable rate-recall tradeoff. Future work will focus on reducing construction cost, and developing more efficient parent-reference strategy to further improve encoding quality.

\section*{Acknowledgments}
This work was supported by the Key R\&D Program of Shandong Province(Grant No. 2023CXPT056), National Natural Science Foundation of China (Grant Nos. 62372350, 62472337, 62220106004, 62232013), Foundation for Innovative Research Groups of the National Natural Science Foundation of China (Grant No.62121001), Fellowship of China National Postdoctoral Program for Innovation Talents (Grant No. BX20230279), Natural Science Basic Research Program of Shaanxi(Grant Nos. 2025JC-JCQN-084, 2025JC-YBQN-792), Basic Research Program (Grant No.JCKY2023110C079).

OpenAI ChatGPT was used to assist in generating portions of the C++ and CUDA implementation. The resulting implementation was reviewed and validated by the authors against an early independently developed Julia implementation. AI tools were also used to a limited extent during the development of the Julia implementation. All code and experimental results were verified by the authors.

{\appendices
\section{Derivations and Implementation Details for STLQ Construction}
\subsection{Derivation of Prefix-LSM}
\subsubsection{Normal Equations and a Prefix Extension}
For the selected prefix $\mathbf{C}_k^r$ in 
\begin{equation}
    \min_{\alpha}(\mathbb{E}_k)= \min_{\alpha}\|\mathbf{x} - \mathbf{C}_k^r\alpha_k\|^2,\quad \alpha_k\in\mathbb{R}^{k\times 1},
    \label{fun_prelsm_coeff_obj}
\end{equation}
let $\mathbf{V}={\mathbf{C}_k^r}^{T}\mathbf{x}$. The least-squares coefficients satisfy
\begin{equation}
  \mathbf{G}_k\alpha_k=\mathbf{V},
  \qquad \mathbf{G}_k={\mathbf{C}_k^r}^{T}\mathbf{C}_k^r.
  \label{fun_ori_normal_prelsm}
\end{equation}
Appending a candidate $\mathbf{c}_n^r$ changes both its own coefficient and those of the prefix:
\begin{equation}
    \min_{\alpha}(\mathbb{E}_{k+1})=\min_{\alpha}\|\mathbf{x} - \mathbf{C}_k^r\tilde{\alpha}_k -\alpha_n\mathbf{c}_n^r \|^2.
\end{equation}
With $\mathbf{U}={\mathbf{C}_k^r}^{T}\mathbf{c}_n^r$, $g=\|\mathbf{c}_n^r\|^2$, and $v={\mathbf{c}_n^r}^{T}\mathbf{x}$, the augmented normal equations are:
\begin{equation}
  \begin{bmatrix}
    \mathbf{G}_k & \mathbf{U}\\
    \mathbf{U}^{T} & g
  \end{bmatrix}
  \begin{bmatrix}\tilde{\alpha}_k\\\alpha_n\end{bmatrix}
  =\begin{bmatrix}\mathbf{V}\\v\end{bmatrix}.
  \label{fun_block_prelsm}
\end{equation}
Subtracting Eq. \ref{fun_ori_normal_prelsm} from the first block row gives:
\begin{equation}
    \tilde{\alpha}_k=\alpha_k - \alpha_n\mathbf{G}_k^{-1}\mathbf{U}.
    \label{fun_solution_prefix_lsm}
\end{equation}
Substituting this into the second block row yields:
\begin{equation}
  \alpha_n = 
  (g-\mathbf{U}^{T}\mathbf{G}_k^{-1}\mathbf{U})^{-1}
  (v-\mathbf{U}^{T}\alpha_k).
\end{equation}

\subsubsection{Incremental energy}
Expanding the prefix error and applying the normal equations gives:
\begin{align}
    \min_{\alpha}(\mathbb{E}_k) 
    &=\|\mathbf{x}\|^2-2\mathbf{V}^T\alpha_k
         +\alpha_k^T\mathbf{G}_k\alpha_k \nonumber\\
    &= \|\mathbf{x}\|^2 - \mathbf{V}^T\alpha_k.
    \label{fun_energy_prelsm_ori}
\end{align}
The corresponding identity for the augmented state is:
\begin{equation}
    \min_{\alpha}(\mathbb{E}_{k+1})= \|\mathbf{x}\|^2 - \mathbf{V}^T\tilde{\alpha}_k -v\alpha_n.
    \label{fun_energy_prelsm_addone}
\end{equation}
Then, combine Eq. \ref{fun_solution_prefix_lsm}, Eq. \ref{fun_energy_prelsm_ori} and Eq. \ref{fun_energy_prelsm_addone} to give:
\begin{equation}
    \min_{\alpha}(\mathbb{E}_{k+1})=\min_{\alpha}(\mathbb{E}_{k}) - \alpha_n(v-\mathbf{U}^{T}\alpha_k).
    \label{fun_energy_prelsm_next}
\end{equation}

\subsection{Derivation of Root Codebook Update}
With assignments and coefficients fixed, the root codebook update is a weighted dimension-wise LSM. We have the normal equation:
\begin{equation}
  [\mathbf{C}_1^r,\mathbf{C}_s^r]\mathbf{A}\mathbf{A}^T
      =\mathbf{X}\mathbf{A}^T,
  \qquad \mathbf{X}\in\mathbb{R}^{d\times N}.
\end{equation}
Define $\mathbf{U}=\mathbf{A}\mathbf{A}^T$ and $\mathbf{V}=\mathbf{X}\mathbf{A}^T$. Partitioning $\mathbf{U}$ according to the first layer and the remaining layers gives:
\begin{equation}
  [\mathbf{C}_1^r,\mathbf{C}_s^r]
  \begin{bmatrix}
    \mathbf{U}_{11} & \mathbf{U}_{1s}\\
    \mathbf{U}_{s1} & \mathbf{U}_{ss}
  \end{bmatrix}
  =[\mathbf{V}_1,\mathbf{V}_s].
\end{equation}
The relevant dimensions are $\mathbf{V}_1\in\mathbb{R}^{d\times h_1^r}$, $\mathbf{V}_s\in\mathbb{R}^{d\times H_s^r}$.

The first block equation is $\mathbf{C}_1^r\mathbf{U}_{11}+\mathbf{C}_s^r\mathbf{U}_{s1}=\mathbf{V}_1$. Hence:
\begin{equation}
  \mathbf{C}_1^r=
    (\mathbf{V}_1-\mathbf{C}_s^r\mathbf{U}_{s1})\mathbf{U}_{11}^{-1}.
  \label{fun_solve_root_fisrt_layer}
\end{equation}
Substituting it into the second block equation gives:
\begin{equation}
  \begin{aligned}
    \mathbf{C}_s^r
      (\mathbf{U}_{ss}-\mathbf{U}_{s1}\mathbf{U}_{11}^{-1}\mathbf{U}_{1s})
      =\mathbf{V}_s-\mathbf{V}_1\mathbf{U}_{11}^{-1}\mathbf{U}_{1s}.
  \end{aligned}
  \label{app:root-schur-equation}
\end{equation}

\subsection{Construct Linked-ADC Norms using Lookup Tables}
The reconstruction of relative residual between a linked node $\mathbf{x}_i$ and its parent $\mathbf{x}_{p(i)}$ is:
\begin{equation}
  \delta_i=\sum_{m=1}^{M}\hat{\alpha}_{i,m}\mathbf{c}^o_{i,m}.
\end{equation}
Expand the norm of reconstructed node $\hat{\mathbf{x}}_i$ as:
\begin{equation}
    \| \hat{\mathbf{x}}_i \|^2 = \| \hat{\mathbf{x}}_{p(i)} \|^2 + 2\hat{\mathbf{x}}_{p(i)}^T\delta_i + \|\delta_i\|^2.
    \label{fun_linked_norm}
\end{equation}
Define a helper function $\Omega(\mathbf{C},t)$ to represent the norm term:
\begin{equation}
    \Omega(\mathbf{C},t) = \sum_{m=t}^{M}\hat{\alpha}_m^2\|\mathbf{c}_{m}\|^2 + 2\sum_{m=t,m<j}^{M}\hat{\alpha}_m\hat{\alpha}_j\mathbf{c}_{m}^{T}\mathbf{c}_{j},
\end{equation}
where $\mathbf{C}$ could be $\mathbf{C}^r$ or $\mathbf{C}^o$. Thus, for root nodes $D(\hat{\mathbf{x}}_l) = 0$, $\|\delta_i\|^2$ is given by:
\begin{equation}
    \|\delta_i\|^2 
    = \hat{\alpha}_1^2\|\mathbf{c}_1^r\|^2 +2\hat{\alpha}_1{\mathbf{c}_1^r}^{T}\sum_{m=2}^{M}\hat{\alpha}_m\mathbf{c}_m^r +  \Omega(\mathbf{C}^r,2).
    \label{fun_norm2_root}
\end{equation}
For linked nodes $D(\hat{\mathbf{x}}_i) \ge 1$, $\|\delta_i\|^2$ is given by:
\begin{equation}
    \|\delta_i\|^2 = \Omega(\mathbf{C}^o,1).
    \label{fun_norm2_linked}
\end{equation}
Then, Eq. \ref{fun_norm2_root}, Eq. \ref{fun_norm2_linked} and the $\hat{\mathbf{x}}_{p(i)}^T\delta_i$ term in Eq. \ref{fun_linked_norm} can be recursively accumulated from ${\mathbf{C}^r}^{T}\mathbf{C}^o$ and ${\mathbf{C}^o}^{T}\mathbf{C}^o$, avoiding the heavy cost of ${\mathbf{C}^r}^{T}\mathbf{C}^r$ when the first root layer is too large.

\section{Residual Codebook Initialization Modes}
The preliminary outward-expanding round produces linked residuals before $\mathbf{C}^o$ is available. As the $h_1^r$ may be large, we design three modes: Legacy, Fast, and Hybrid to trade offline cost against initialization quality.

\subsubsection{Legacy}
The preliminary topology is built using the entire $\mathbf{C}^r$. This normally provides the best recall in our experiments, but ILS and ICM can be expensive when the integrated IVF layer has a large $h_1^r$.

\subsubsection{Fast}
Fast mode aggregates the $h_1^r$ codewords of $\mathbf{C}_1^r$ into a smaller set of size $h$ using spherical k-means. This reduces the codeword search cost of the preliminary round.

\subsubsection{Hybrid}
Hybrid mode runs several rounds of Legacy mode, and then switches to Fast mode. Although the recall of Fast and Hybrid modes decreases moderately, they save offline training time. Nevertheless, neither mode consistently achieves better recall than the other.

}

\bibliographystyle{IEEEtran}
\bibliography{IEEEabrv, ./bib/ref.bib}


 




\vfill

\end{document}